\documentclass[10pt,twocolumn,letterpaper]{article}

\usepackage{iccv}              
\usepackage[accsupp]{axessibility}

\definecolor{iccvblue}{rgb}{0.21,0.49,0.74}
\usepackage[pagebackref,breaklinks,colorlinks,allcolors=iccvblue]{hyperref}
\usepackage{booktabs}
\usepackage{float}
\usepackage[dvipsnames]{xcolor}
\usepackage{amsmath,array}
\usepackage{colortbl} 
\definecolor{gg}{HTML}{e2f0cb}
\definecolor{bl}{HTML}{5E91D1}
\definecolor{gr}{HTML}{7DBE65}
\definecolor{rd}{HTML}{F2AA84}
\usepackage{makecell}
\usepackage{multirow}
\usepackage{algpseudocode}
\usepackage{textcomp}
\usepackage{csquotes}
\usepackage{algorithm}
\usepackage{algpseudocode}

\def\paperID{****} 
\def\confName{ICCV}
\def\confYear{2025}

\title{Perspective-aware 3D Gaussian Inpainting with Multi-view Consistency}
\author{Yuxin Cheng, Binxiao Huang, Taiqiang Wu, Wenyong Zhou, Chenchen Ding \\
Zhengwu Liu, Graziano Chesi, Ngai Wong\thanks{Corresponding author} \\
The University of Hong Kong, Hong Kong SAR, China\\
{\tt\small \{yxcheng,huangbx7,takiwu,wenyongz,dingcc\}@connect.hku.hk, \{zwliu,chesi,nwong\}@eee.hku.hk}
}
\begin{document}
\maketitle
\begin{abstract}
3D Gaussian inpainting, a critical technique for numerous applications in virtual reality and multimedia, has made significant progress with pretrained diffusion models. However, ensuring multi-view consistency, an essential requirement for high-quality inpainting, remains a key challenge. In this work, we present PAInpainter, a novel approach designed to advance 3D Gaussian inpainting by leveraging perspective-aware content propagation and consistency verification across multi-view inpainted images. Our method iteratively refines inpainting and optimizes the 3D Gaussian representation with multiple views adaptively sampled from a perspective graph. By propagating inpainted images as prior information and verifying consistency across neighboring views, PAInpainter substantially enhances global consistency and texture fidelity in restored 3D scenes. Extensive experiments demonstrate the superiority of PAInpainter over existing methods. Our approach achieves superior 3D inpainting quality, with PSNR scores of 26.03 dB and 29.51 dB on the SPIn-NeRF and NeRFiller datasets, respectively, highlighting its effectiveness and generalization capability. The code will be publicly available at \href{https://pa-inpainter.github.io/}{https://pa-inpainter.github.io}.
\end{abstract}    
\section{Introduction}
\label{sec:intro}

As a prominent application in the realm of 3D editing, 3D inpainting plays a pivotal role in various applications and industries, including the metaverse and holographic multimedia production~\cite{zhao2022metaverse}. However, traditional hand-crafted 3D completion approaches, which rely on professional designers and specialized tools, remain labor-intensive and cumbersome. With recent advancements in 3D neural representations~\cite{kerbl3Dgaussians, cheng2025re, charatan2024pixelsplat, lu2024scaffold} and generative models~\cite{Rombach2022SD2}, 3D inpainting can be achieved by applying a two-stage paradigm: 1) using a pretrained 2D diffusion model to inpaint masked multi-view renderings of the 3D Gaussian scene with missing regions; and 2) optimizing the initial 3D Gaussian scene with the inpainted multi-view images~\cite{prabhu2023inpaint3d, cao2024mvinpainter}. While this efficient framework shows significant potential, multi-view inconsistency remains an inherent challenge in diffusion models due to their independent view processing nature, which hinders high-quality 3D inpainting~\cite{weber2024nerfiller, liu2024infusion}.

\begin{figure}[t]
  \centering
   \includegraphics[width=\linewidth]{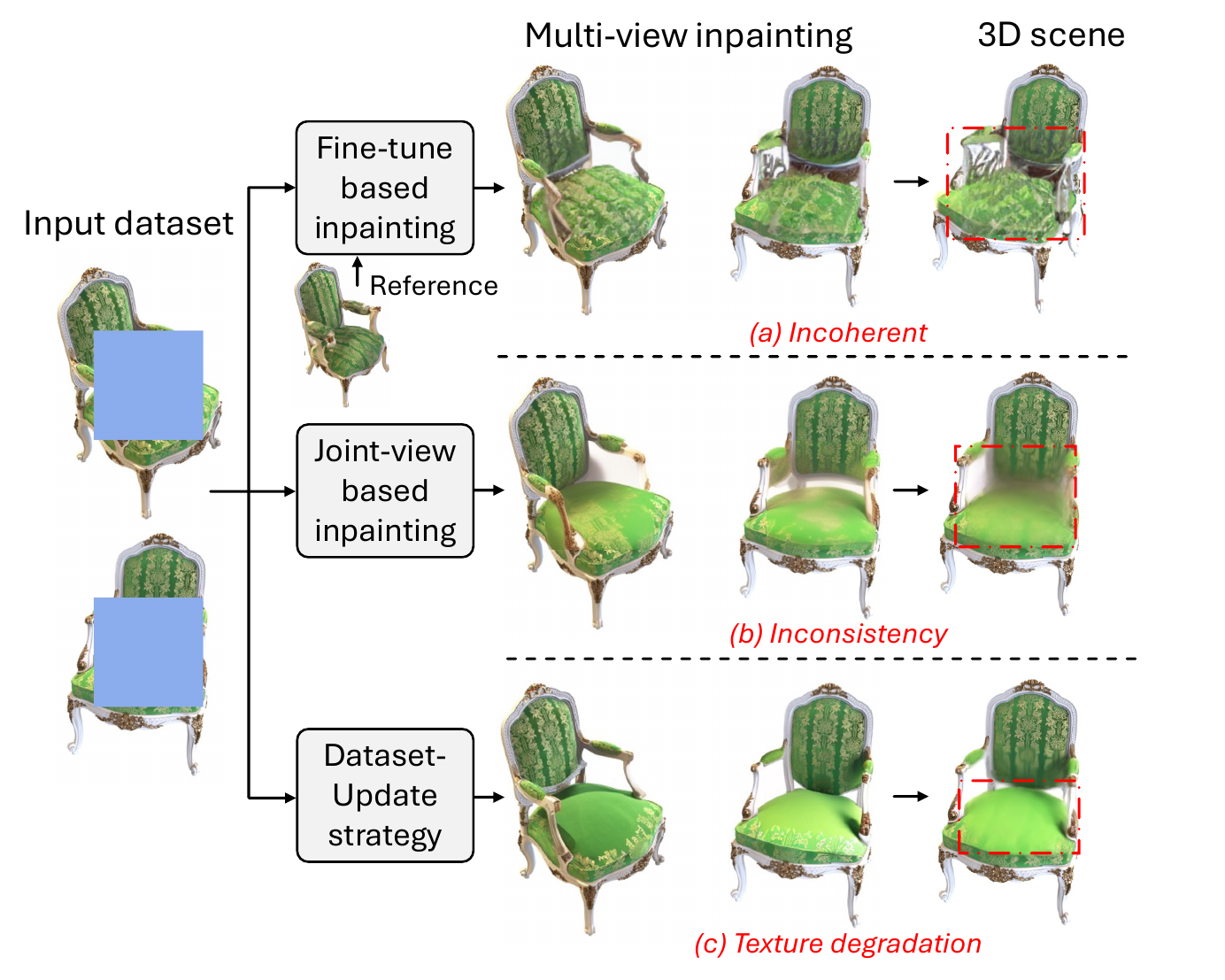}
   \caption{Current challenges in 3D Gaussian inpainting: a) the fine-tune based inpainter trained for specific tasks~\cite{cao2024mvinpainter} experiences significant performance decline when applied to general inpainting scenarios; b) the joint-view inpainting method~\cite{weber2024nerfiller} struggles with inconsistency across multi-view images, resulting in noisy inpainting results; c) the DU strategy~\cite{haque2023instruct} leads to texture degradation in both inpainted multi-view images and the final 3D scene. }
   \label{fig:motivation}
\end{figure}

Existing works have explored various approaches to improve multi-view consistency in 3D Gaussian inpainting, yet new limitations continue to emerge. Fine-tune based inpainting methods adapt diffusion models with additional control conditions (e.g., reference images~\cite{cao2024mvinpainter}), but are confined to specific scenarios, as illustrated in \cref{fig:motivation}(a). Without modifying pretrained diffusion models, the joint-view based inpainting approach~\cite{weber2024nerfiller} processes multi-view images in 2×2 grid tiles and achieves improved consistency, yet still exhibits artifacts in challenging regions, as shown in \cref{fig:motivation}(b). Similarly, DatasetUpdate (DU)~\cite{haque2023instruct} alternates between 3D scene optimization and multi-view inpainting while progressively updating the dataset to improve consistency. However, it suffers from texture fading in the final results, as demonstrated in \cref{fig:motivation}(c). These limitations highlight the persistent challenge of achieving high-fidelity and globally consistent inpainting across multiple views.

\begin{figure}[t]
  \centering
   \includegraphics[width=\linewidth]{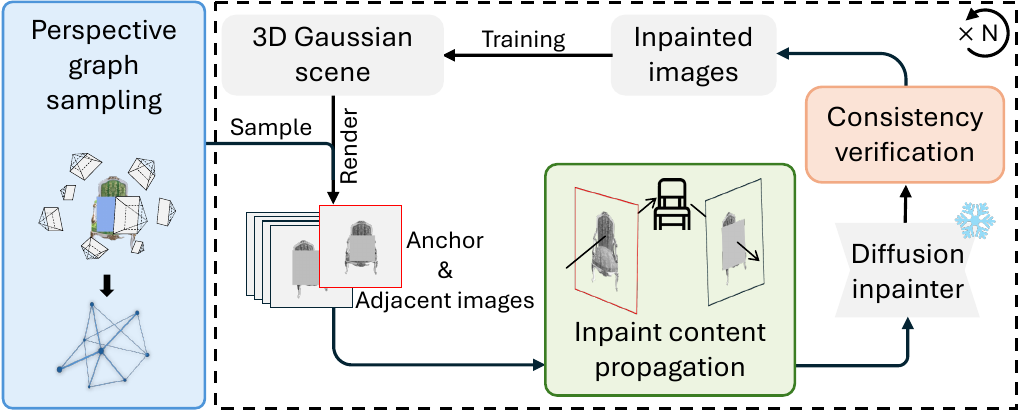}
   \caption{The overall pipeline of our proposed PAInpainter. Based on the constructed perspective graph, our approach iteratively performs multi-view image inpainting and 3D Gaussian training. The adaptive graph sampling algorithm enables efficient inpaint content propagation across adjacent viewpoints, while consistency verification ensures coherent multi-view inpainting results, thereby improving the 3D inpainting quality.}
   \label{fig:highlevel}
\end{figure}

In this paper, we introduce the \textbf{P}erspective-\textbf{A}ware 3D Gaussian \textbf{Inpainter} (PAInpainter) to enhance multi-view consistency. As illustrated in \cref{fig:highlevel}, we propose a novel perspective-aware framework with three key components: perspective graph sampling, inpaint content propagation, and consistency verification. Specifically, we construct a perspective graph that models the spatial relationships among viewpoints. Leveraging adaptive graph sampling and the inherent perspective overlap between neighboring views, we propagate inpainted content across adjacent cameras, which serves as supplementary visual priors for the diffusion model during inpainting, improving fine-grained texture preservation and consistency across multi-view inpainted images. To ensure high-quality and reliable results, we introduce a dual-feature verification mechanism that evaluates both texture and geometric coherence in latent space, effectively identifying and selecting consistent inpainting results. Combined with our framework, the versatile generation capability of the pretrained diffusion model further empowers our approach to handle various challenging 3D inpainting scenarios.

Our approach demonstrates exceptional performance in high-fidelity 3D Gaussian inpainting across diverse scenarios. Through extensive experiments on three mainstream 3D inpainting datasets, we demonstrate that PAInpainter significantly outperforms existing methods both quantitatively and qualitatively. Additionally, our PAInpainter exhibits robust generalization capability across various scenarios. Our main contributions are summarized as follows:
\begin{itemize}
    \item \textcolor{black}{We propose a novel perspective-aware framework for 3D Gaussian inpainting that systematically integrates inpainting view sampling, cross-view content propagation, and consistency verification.}
    \item We introduce three effective components: a perspective graph to guide viewpoint sampling for inpainting, a perspective-aware projection strategy to propagate inpainting content, and a dual-feature verification mechanism to ensure multi-view consistency.
    \item Extensive experiments on diverse 3D scenes demonstrate that PAInpainter outperforms state-of-the-art methods in achieving superior consistency and visual fidelity.
\end{itemize}
\section{Related Work}
\label{sec:relat}

\subsection{2D Image Inpainting}
2D inpainting methods aim to restore missing or obscured regions in images with coherent textures and structures~\cite{paragios2006handbook, Bertalmio_Sapiro_Caselles_Ballester_2000}. Early approaches relied on texture synthesis and pixel interpolation techniques by leverages information from known regions~\cite{Efros_Leung_1999, Barnes_Shechtman_Finkelstein_Goldman_2009, Darabi_Shechtman_Barnes_Goldman_Sen_2012}. Learning-based approaches, especially deep learning methods~\cite{Ko_Kim, Yu_Lin_Yang_Shen_Lu_Huang_2018, pathak2016context, li2022mat, yu2023inpaint} and recent diffusion models~\cite{Rombach2022SD2}, have since emerged as powerful alternatives, demonstrating superior capabilities in high-fidelity content completion. The Latent Diffusion Models (LDMs)\cite{Rombach2022SD2} and its variants\cite{lugmayr2022repaint, zhang2023control} achieve remarkable generation results across diverse scenarios. However, these methods process each image independently without considering 3D spatial relationships and geometric attributes among multiple viewpoints, leading to subsequent inconsistency when applied to the 3D domain.

\begin{figure*}[t]
  \centering
   \includegraphics[width=\linewidth]{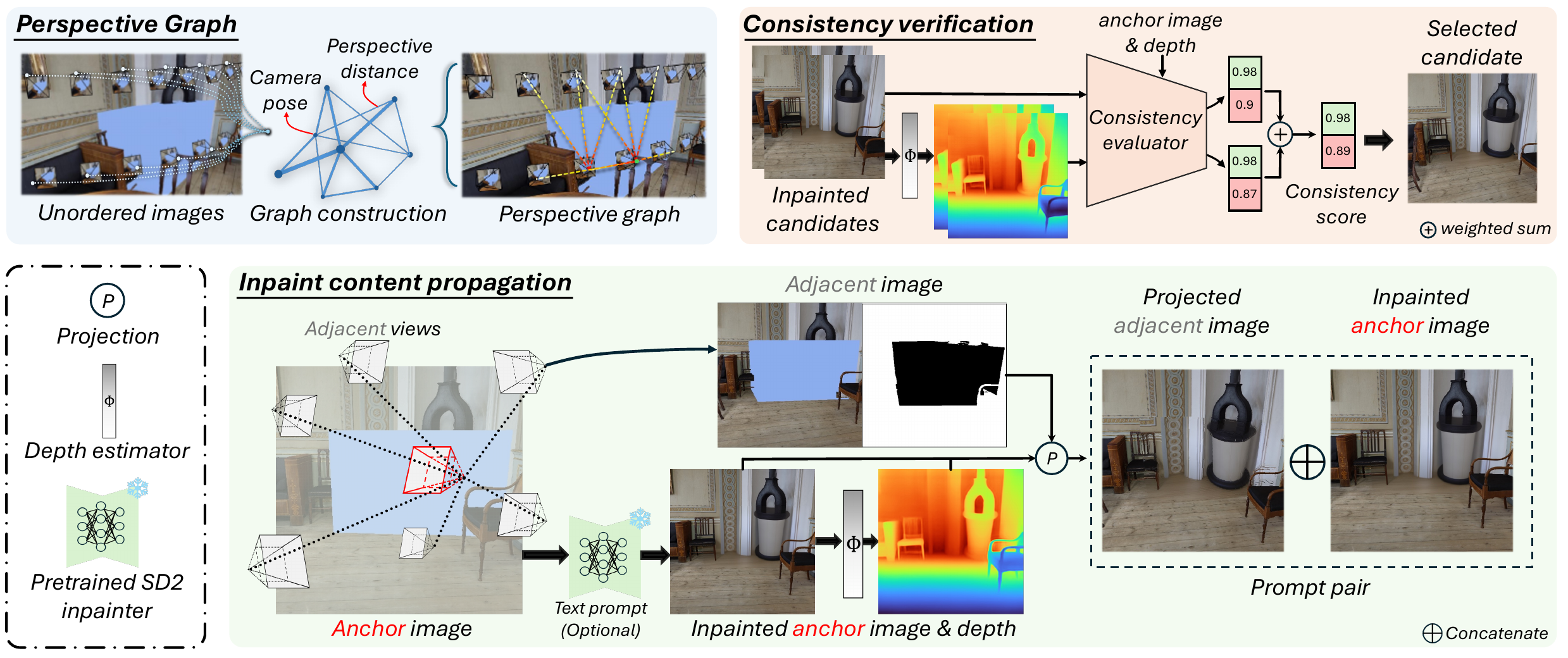}
   \caption{Overview of PAInpainter for multi-view consistent 3D Gaussian inpainting. Our method is built upon the pretrained SD2~\cite{Rombach2022SD2} and incorporates three key components: 1) \textit{\textcolor{bl}{perspective graph}} models spatial relationships among cameras to guide adjacent view sampling; 2) \textit{\textcolor{gr}{inpaint content propagation}} transmits inpainting content across adjacent views sampled from perspective graph, providing extra visual priors for diffusion inpainting; 3) \textit{\textcolor{rd}{consistency verification}} evaluates inpainted results based on texture and geometric features coherence. The perspective-aware graph sampling contributes to effective content propagation and consistency verification across multiple views.}
   \label{fig:pipeline}
\end{figure*}

\subsection{3D Scene Inpainting}
3D inpainting extends the content completion task into 3D space. Early approaches focused on geometric completion using traditional representations like point clouds and meshes~\cite{dai2017shape, wu2018shapehd}. Recent advances in neural representations, particularly Neural Radiance Fields (NeRF)\cite{mildenhall2021nerf} and 3D Gaussian Splatting (3DGS)\cite{kerbl3Dgaussians}, have revolutionized 3D scene modeling. While direct 3D diffusion models~\cite{muller2023diffrf, shue20233d, warburg2023nerfbusters} face challenges with limited training data and computational complexity, an alternative approach combines pretrained 2D diffusion models with 3D neural representations~\cite{jiang2023inpaint4dnerf, mirzaei2023spin, liu2022nerf, weber2024nerfiller, prabhu2023inpaint3d, lin2024taming, lu2024view, wang2024innerf360, weder2023removing}. This paradigm shows promising results by combining the powerful generation capabilities of 2D diffusion models with real-time 3D reconstruction~\cite{cao2024mvinpainter, liu2024infusion}. Despite the advancement in 3D inpainting efficiency, ensuring geometric and appearance consistency across different viewpoints remains challenging. In this paper, we build our method on 3DGS, which enables fast training and real-time rendering, and achieve improved multi-view consistency by propagating extra prior information to guide the diffusion inpainting process.

\subsection{Multi-view Consistency}
Multi-view consistency ensures that the generated content in multi-view images of a 3D scene maintains geometric and texture coherence~\cite{haque2023instruct, watson2022novel}. Recent works have explored two main techniques to address inconsistency arising from 2D diffusion models. The first approach resorts to the diffusion model fine-tuning with additional conditions~\cite{watson2022novel, kant2024spad, muller2024multidiff, wu2024reconfusion}, namely incorporating depth features, task-specific modules, and geometric constraints~\cite{liu2024infusion, cao2024mvinpainter, cao2024leftrefill}. However, these methods typically specialize in specific tasks like object removal and struggle to generalize to broader 3D inpainting scenarios. The second technique explores solution without modifying pretrained models~\cite{haque2023instruct, jiang2023inpaint4dnerf, weber2024nerfiller}, but leverages depth priors or additional supervision~\cite{prabhu2023inpaint3d, spinnerf} to enhance cross-view consistency in generation and reconstruction process. While these approaches show potential, they often lack proactive consistency inspection of inpainted images, leading to compromised performance under challenging conditions. To address this limitation, we introduce a dual-feature verification mechanism designed to reject inconsistencies, thereby ensuring coherent inpainting across diverse scenarios for 3D scene restoration.

\section{Methodology}
\label{sec:metho}
\textbf{Overview.} The key components of PAInpainter are illustrated in \cref{fig:pipeline}. This section first introduces the overall framework (\cref{sec:framework}), followed by the technical details of perspective graph construction, inpaint content propagation, and consistency verification (\cref{sec:PAInpainter}). The adaptive graph sampling strategy and 3D Gaussian training procedure are elaborated in \cref{sec:adaptive}.

\subsection{Framework}
\label{sec:framework}

As shown in \cref{fig:highlevel}, PAInpainter completes unknown regions within a 3D scene by iteratively inpainting multi-view renderings and optimizing the 3D Gaussians with inpainted images. Building upon the high-fidelity image inpainting capabilities of pretrained StableDiffusion2 (SD2)~\cite{Rombach2022SD2}, our framework enhances the multi-view inpainting consistency through three key techniques: perspective graph sampling, inpaint content propagation and consistency verification.

Based on the fact that the cross-attention mechanism of SD2 allows for reference-guided content generation in missing regions~\cite{cao2024mvinpainter}, we observe that the inpainting consistency significantly degrades with increasing perspective differences between views, as shown in \cref{fig:motivation}(a). This observation motivates us to sample images with similar perspectives, thereby promoting the generation of consistent content. These adjacent views serve dual purposes: they facilitate effective content propagation by providing reliable texture and geometric priors for masked images, while enabling feature-space consistency verification among inpainting images to mitigate SD2's inherent randomness and select optimal results from multiple candidates.

Based on these findings, we develop PAInpainter based on the following iterative framework:

\begin{enumerate}
    \item Given a 3D Gaussian scene $\mathcal{G}_u$ with unknown regions, multi-view images $\mathbf{I}=\{I_i\}_{i=1}^n$ with corresponding camera poses $\mathbf{T} = \{T_i \in \mathrm{SE}(3)\}_{i=1}^n$ and masks $\mathbf{M}=\{M_i\}_{i=1}^n$, we construct a perspective graph $\mathbf{G}$ for $\mathbf{I}$, where edges encode the perspective distances among views.
    \item For each inpainting round, we adaptively sample an anchor image $I_{anchor}$ from the constructed graph and employ SD2 to inpaint it, obtaining $I_{anchor}'$. We then query its adjacent images from $\mathbf{G}$ to form a subset $\mathbf{I}_{adj}$. The inpainted content from $I_{anchor}'$ is projected to each image in $\mathbf{I}_{adj}$, and $I_{anchor}'$ serves as a reference image for following diffusion inpainting of these adjacent views.
    \item For images in $\mathbf{I}_{adj}$, multiple inpainted candidates are generated by SD2. We then compute consistency scores between each candidate and $I_{anchor}'$ with regard to the inpainting regions, selecting the candidate with the highest score as the final inpainting result.
    \item We optimize the 3D Gaussian scene $\mathcal{G}_u$ by training on the inpainted images.
\end{enumerate}
\noindent
The process iteratively alternates between multi-view inpainting (steps 2-3) and 3D Gaussian optimization (step 4), progressively inpainting and refining the 3D scene.

\begin{figure}[t]
    \centering
    \includegraphics[width=\linewidth]{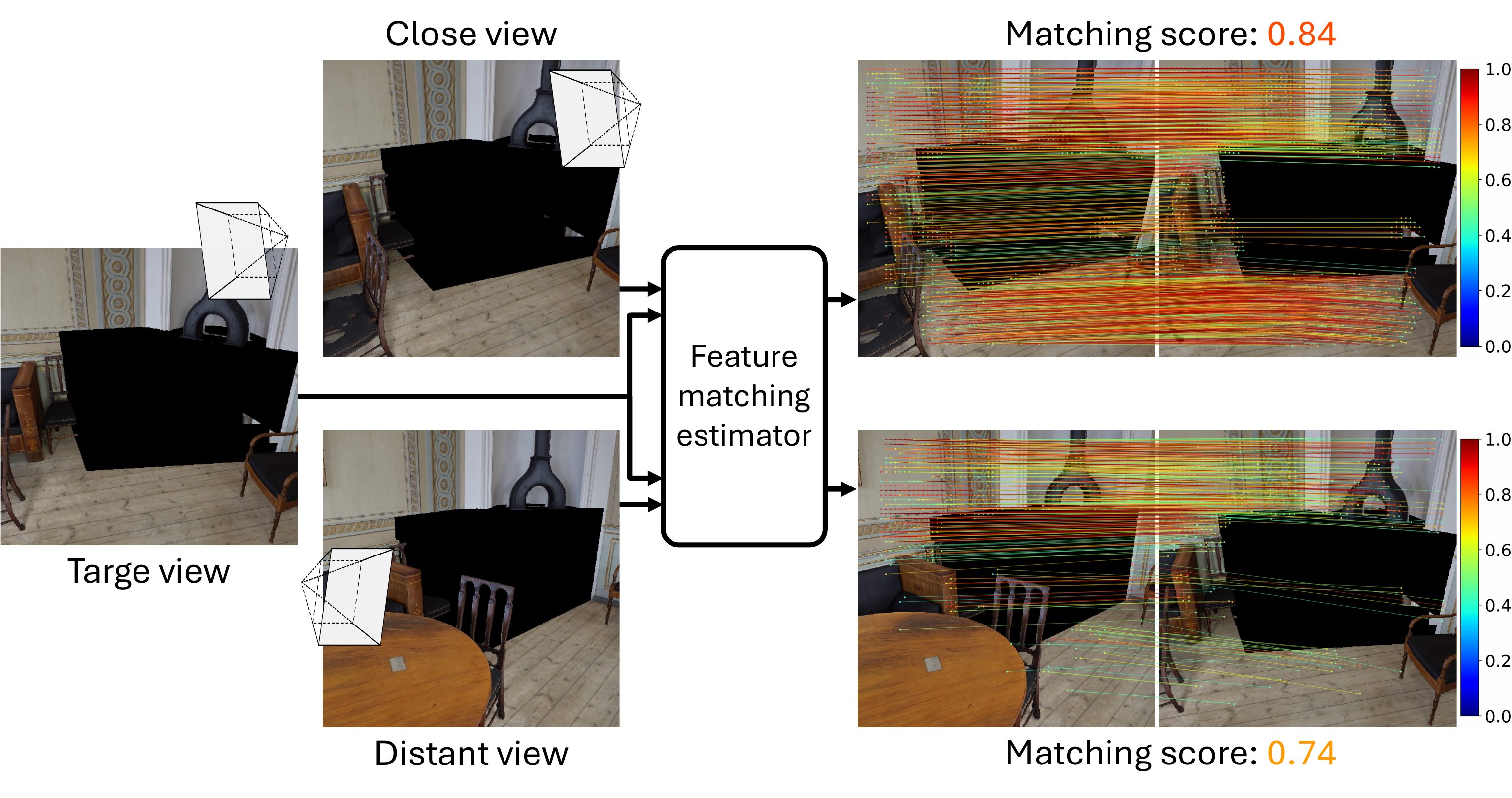}
    \caption{Perspective distance evaluation via feature matching. The color bar indicates the match's confidence. For a given target image, nearby views achieve significantly higher matching score (0.84) compared to distant views (0.74), validating the effectiveness of our perspective-aware graph construction method. This distance metric naturally captures the spatial relationships between different viewpoints.}
    \label{fig:graph}
\end{figure}

\subsection{PAInpainter}
\label{sec:PAInpainter}
We now detail the three key modules of PAInpainter for achieving consistent 3D Gaussian inpainting.

\noindent
\textbf{Perspective graph construction.} The graph $\mathbf{G}$ underpins our entire inpainting pipeline by modeling the proximity relationships among diverse viewpoints. \textcolor{black}{Although the cameras' poses are available, the view difference, i.e., the captured content in the  images, cannot be solely described by the 6-DoF distance due to variations in perspective, orientation, and scene geometry~\cite{Wang_2014_CVPR, arandjelovic2016netvlad}. To solve this problem,} we propose evaluating view similarities through feature matching metrics, as shown in \cref{fig:graph}. Specifically, we employ LoFTR~\cite{sun2021loftr} for its transformer-based architecture that enables robust feature matching under challenging viewpoint changes. For image pair $(I_i, I_j)$ in the dataset, we extract matches with confidence scores above threshold $\tau$ ($\tau=0.4$ fixed in our implementation). The perspective distance is evaluated based on the average confidence score for these matches, where a higher average matching score indicates a smaller distance. In the final graph $\mathbf{G}$, nodes store images with their camera poses and masks, while edges encode the computed perspective distances. This perspective-aware graph enables effective sampling of adjacent views for consistent inpainting. As demonstrated by our experiments, this strategy provides enhanced robustness to viewpoint variations while preserving geometric interpretability.

\begin{figure}
    \centering
    \includegraphics[width=\linewidth]{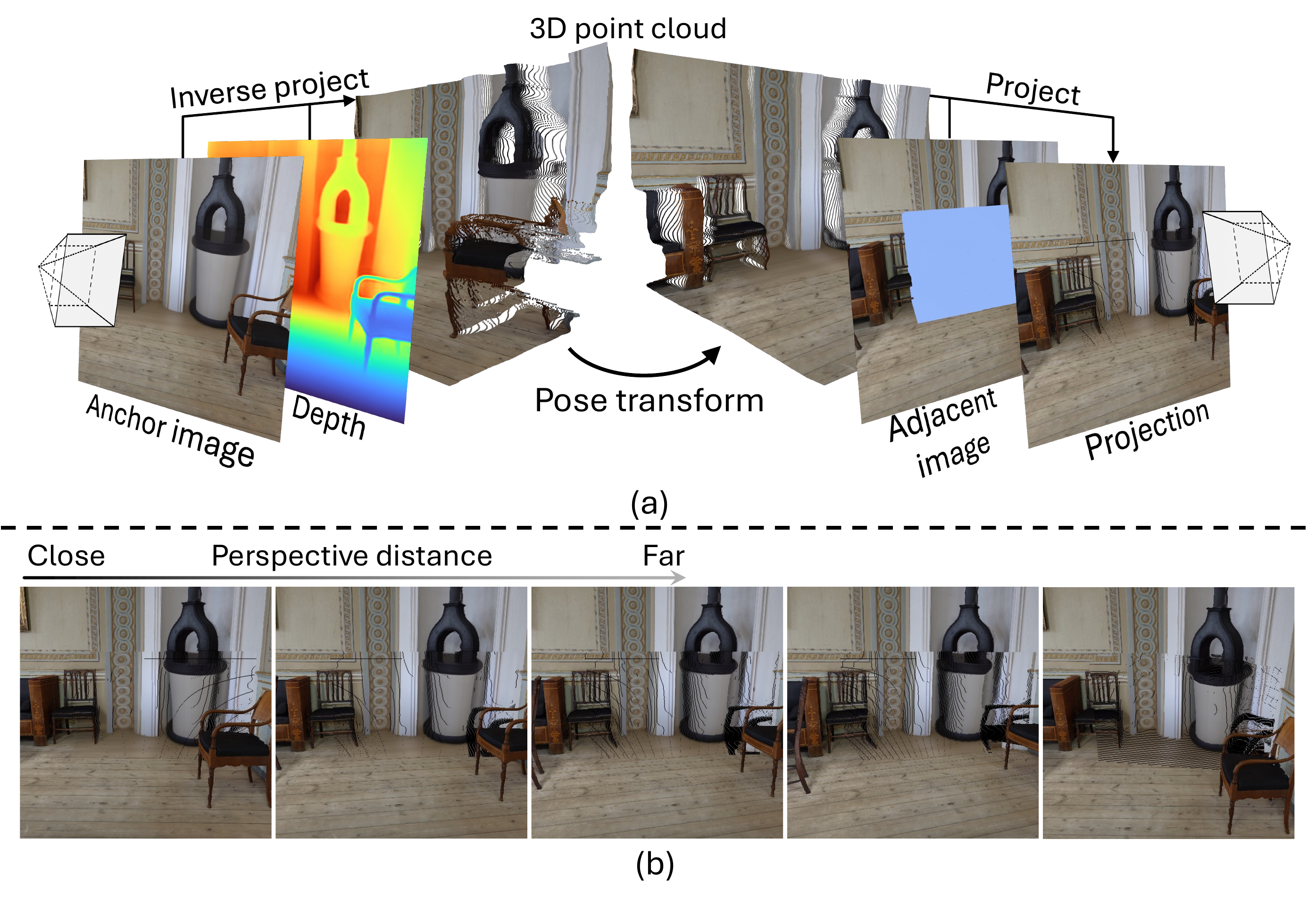}
    \caption{Inpaint content propagation mechanism. (a) Using depth information and camera poses, inpainted content from an anchor image is projected onto its adjacent masked views. (b)  The projection results on neighboring views sampled from our perspective graph. Thanks to our graph-based sampling strategy, most masked regions in the selected images receive ample projected content (high coverage of effective pixels), \textcolor{black}{offering rich prior information for subsequent SD2 inpainting.}
    }
    \label{fig:projection}
\end{figure}

\noindent
\textbf{Inpaint content propagation.} \textcolor{black}{To enhance multi-view consistency and high-fidelity inpainting, we feed supplementary priors of masked region along with the image to SD2 via inpaint content propagation.} Guided by camera poses sampled from graph $\mathbf{G}$, we render the anchor image $I_{anchor}$ and its top-$k$ adjacent images $\mathbf{I}_{adj}=\{I_i^{adj}\}_{i=1}^k$ from 3D Gaussian scene $\mathcal{G}_u$. We first independently inpaint the anchor image $I_{anchor}$ using SD2, obtaining $I_{anchor}'$, followed by propagating the $I_{anchor}'$ to its adjacent images $\mathbf{I}_{adj}$ with masked regions through perspective projection to offer extra prior for SD2 inpainting, as shown in \cref{fig:projection} (a).

Specifically, we use ZoeDepth~\cite{bhat2023zoedepth} to estimate the depth map $d_{anchor}$ for $I_{anchor}'$. With $d_{anchor}$ and camera parameters (intrinsic $K$ and extrinsic $T_{anchor}$), we inversely project the 2D image $I_{anchor}'$ into perspective coordinates by
\begin{equation}
     \begin{bmatrix} x_c, y_c, z_c \end{bmatrix}^\top = K^{-1} \cdot (\begin{bmatrix} u, v, 1\end{bmatrix}^\top \cdot d),  
\end{equation}
where $[u, v, 1]^\top$ and $d$ represent 2D image coordinate and depth value, respectively, and $[x_c, y_c, z_c]^\top$ represents the 3D coordinate. For each adjacent image $I_i^{adj}$ with camera pose $T_i$, we transform the 3D point cloud from anchor perspective to the perspective coordinates of $I_i^{adj}$ by
\begin{equation}
     \begin{bmatrix} x_c', y_c', z_c', 1 \end{bmatrix}^\top = T_i \cdot T_{anchor}^{-1} \cdot \begin{bmatrix} x_c, y_c, z_c, 1 \end{bmatrix}^\top, 
\end{equation}
obtaining $[x_c', y_c', z_c']^\top$ after homogeneous normalization. We project these coordinates onto $I_i^{adj}$, updating only pixels within the masked region. \textcolor{black}{For regions where projection fails due to view differences or depth estimation errors, we retain the rendering RGB values from 3D Gaussian scene.}

Leveraging our perspective graph sampling strategy, the projection effectively propagates the inpainted content from anchor image to adjacent frames while preserving geometric and texture consistency, as shown in \cref{fig:projection} (b). The propagated adjacent images $\mathbf{I}_{adj}$ are then paired with $I_{anchor}'$ as reference guidance for SD2 diffusion inpainting.

\noindent
\textbf{Consistency verification.} Obstacles arising from perspective differences make it impossible for consistency verification to rely solely on pixel comparison. Therefore, we elevate the consistency verification process into feature space. We independently generate multiple inpainting candidates for each masked adjacent image. To handle varying 3D inpainting scenarios, we propose verifying consistency by assessing coherence in texture and geometry feature spaces. As shown in \cref{fig:pipeline}, for the inpainted candidates of $I_i^{adj}$, we use ZoeDepth to estimate the corresponding depth maps. We apply a feature extraction model as consistency evaluator (ResNet-18~\cite{he2016resnet}) to separately extract both RGB and depth features for each candidate, as well as for the inpainted anchor image and its depth map. Finally, we compute the cosine similarity between candidates and the inpainted anchor image based on fused dual features. The overall consistency score is computed as a weighted combination of RGB and depth similarities: $S = \eta S_{rgb} + (1-\eta) S_{depth}$, where $\eta$ controls the relative importance of texture and geometry consistency. The candidate with the highest consistency score is then selected as the final inpainting result. The weighting factor $\eta$ is  empirically set to $0.7$ to balance fine texture details and structural coherence and four candidates generated for each image.

Given the small perspective differences between adjacent images and our dual-feature coherence approach, our consistency verification mechanism effectively identifies and excludes inconsistent inpainting results. Specifically, by leveraging hierarchical feature extraction capability of ResNet-18 at multiple scales and the complementary nature of RGB-depth feature pairs, this method significantly enhances the multi-view consistency of images fed into the 3D Gaussian optimization process, thereby improving the overall quality of 3D Gaussian inpainting.

\subsection{Adaptive Sampling \& 3D Gaussian Training}
\label{sec:adaptive}
\noindent
\textbf{Adaptive sampling.} During the iterative process, we strategically sample an anchor image and its $k$ nearest neighbors on our proposed perspective graph $\mathbf{G}$ for inpainting and 3D Gaussian training. In the first iteration, the anchor image is selected from the entire dataset to initialize the process. For each subsequent iteration, we adopt a distance-aware sampling strategy: the anchor image is sampled from the pool of previously inpainted images, excluding both previously selected anchor images and their $k/2$ nearest neighbors from future anchor image selection. Here, $k$ is scene-dependent, determining both the number of adjacent views for inpainting and the spatial separation between anchor images. This spatial constraint ensures well-distributed scene coverage and mitigates the risk of local region trapping. We further maintain a priority queue based on consistency scores from previous iterations, prioritizing images with worse coherence for refinement \textcolor{black}{(the algorithm flowchart in Appendix~B.3)}. This adaptive mechanism progressively improves both global consistency and local detail quality.

\noindent
\textbf{3D Gaussian Training.} Given the dataset consisting of inpainted and masked multi-view images, we optimize 3D Gaussians following the vanilla 3DGS framework~\cite{kerbl3Dgaussians}. The optimization objective combines L1 and D-SSIM losses:
\begin{equation}
    \mathcal{L} = (1-\lambda)\mathcal{L}_1(I_i' - I_i) + \lambda \mathcal{L}_{SSIM}(I_i' - I_i), \ \lambda=0.2,
\end{equation}
where $I_i'$ and $I_i$ denote the rendering and inpainted images respectively. For masked images, we exclude the missing regions from loss computation during optimization.

Our method achieves high-quality multi-view consistent inpainting without fine-tuning pretrained diffusion models. As demonstrated in \cref{fig:titlefigure}, PAInpainter effectively handles diverse 3D inpainting scenarios.
\section{Experiments}
\label{sec:exper}

\begin{table*}[t]
    \centering
    \hspace{-0.25cm}
    \resizebox{1.01\linewidth}{!}{
        \setlength{\tabcolsep}{2pt}
        \begin{tabular}{l|c c c c | c c c c | c c c c }
            \toprule[1.4pt]
                & \multicolumn{4}{c|}{NeRF Blender~\cite{mildenhall2021nerf}} & \multicolumn{4}{c|}{SPIn-NeRF~\cite{spinnerf}} & \multicolumn{4}{c}{NeRFiller~\cite{weber2024nerfiller}} \\ \midrule[1pt]
                & PSNR (dB) $\uparrow$ & SSIM $\uparrow$ & LPIPS $\downarrow$ & FID $\downarrow$ & PSNR (dB) $\uparrow$ & SSIM $\uparrow$ & LPIPS $\downarrow$ & FID $\downarrow$ & PSNR (dB) $\uparrow$ & SSIM $\uparrow$ & LPIPS $\downarrow$ & FID $\downarrow$ \\ \midrule[1pt]
                Masked 3DGS & 11.57 & 0.83 & 0.19 & - &13.46 & 0.41 & 0.40 & - & 12.95 & 0.76 & 0.28 & -  \\
                SD2~\cite{Rombach2022SD2} & 20.42 & 0.90 & 0.09 & \cellcolor{orange!40} 102.2 & 23.48 & 0.73 & 0.23 & \cellcolor{orange!40} 140.3 & 20.36 & 0.84 & 0.17 & \cellcolor{orange!40} 105.2 \\
                MVInpainter~\cite{cao2024mvinpainter} & 19.42 & 0.81 & 0.17 & 148.6 & 24.80 & 0.74 & 0.21 & 152.2 & 21.13 & 0.80 & 0.18 & 117.7 \\
                GridPrior + DU $^\star$ ~\cite{weber2024nerfiller} & \cellcolor{yellow!40} 22.77 & \cellcolor{yellow!40} 0.92 & \cellcolor{orange!40} 0.08 & \cellcolor{yellow!40} 104.2 & \cellcolor{yellow!40} 25.19 & \cellcolor{yellow!40} 0.79 & \cellcolor{yellow!40} 0.20 & 151.2 & \cellcolor{orange!40} 26.97 & \cellcolor{orange!40} 0.92 & \cellcolor{orange!40} 0.13 & 121.9\\
                NeRFiller $^\star$~\cite{weber2024nerfiller} & \cellcolor{orange!40} 23.27 & \cellcolor{orange!40} 0.92 & \cellcolor{yellow!40} 0.09 & 153.7 & \cellcolor{orange!40} 25.20 & \cellcolor{orange!40} 0.79 & \cellcolor{orange!40} 0.17 & \cellcolor{yellow!40} 146.1 & \cellcolor{yellow!40} 22.35 & \cellcolor{yellow!40} 0.88 & \cellcolor{yellow!40} 0.15 & \cellcolor{yellow!40} 110.4\\
                PAInpainter (ours) & \cellcolor{red!40} 24.19 & \cellcolor{red!40} 0.92 & \cellcolor{red!40} 0.08 & \cellcolor{red!40}101.8 & \cellcolor{red!40}26.03 & \cellcolor{red!40}0.81 & \cellcolor{red!40}0.15 & \cellcolor{red!40}121.7 & \cellcolor{red!40}29.51 & \cellcolor{red!40}0.94 & \cellcolor{red!40}0.08 & \cellcolor{red!40}96.1 \\
            \bottomrule[1.1pt]
        \end{tabular}
    }
\caption{\textcolor{black}{Quantitative comparison on the multiple datasets. We compare our method against advanced approaches on three datasets. Higher PSNR and SSIM, as well as lower LPIPS and FID indicate better performance. Cells are highlighted as follows: \colorbox{red!40}{best}, \colorbox{orange!40}{second best}, \colorbox{yellow!40}{third best}. Our method surpasses \textit{all baselines} across these metrics, demonstrating its efficacy and robust generalization. $^\star$ represents replacing the original NeRF backbone with 3DGS for fair comparison. Detailed performance of each scene are provided in the Appendix~C.}}
    \label{tab:quantitative_statistic}
\end{table*}

In our experimental evaluation, we conduct comprehensive comparisons between PAInpainter and state-of-the-art approaches across 28 scenes, spanning 4 distinct inpainting tasks. Our framework employs the pretrained StableDiffusion2 (SD2)~\cite{Rombach2022SD2} as the backbone for image inpainting and ZoeDepth~\cite{bhat2023zoedepth} for monocular depth estimation. Detailed implementation specifics and explanations of hyperparameters are provided in Appendix~B.1 and Appendix~B.2.

\noindent
\textcolor{black}{\textbf{Datasets.} To rigorously evaluate the effectiveness and generalization capability of PAInpainter, we utilize three mainstream datasets. The first dataset consists of 8 object-centric scenes derived from the \textit{NeRF Blender} dataset~\cite{mildenhall2021nerf}, with multi-view images at a resolution of $512\times512$. Missing regions are generated by masking the central $192\times192$ pixels in each image (as exemplified by the \enquote{chair} scene in \cref{fig:titlefigure}). Additionally, we use the \textit{SPIn-NeRF} dataset~\cite{spinnerf} as the second dataset for 3D unbounded scene inpainting. Since the SPIn-NeRF dataset only covers a single 3D inpainting task (foreground object removal), we also incorporate the dataset introduced by \textit{NeRFiller}~\cite{weber2024nerfiller}, which includes 10 real-world complex 3D inpainting scenes. Collectively, our experimental corpus of 28 scenes (details in Appendix~C) encompasses multiple 3D inpainting scenarios: 1) large indoor missing region, 2) object-centric large missing region, 3) object-centric removal, and 4) multiple disjoint missing regions (illustrated in \cref{fig:titlefigure}).}

\noindent
\textbf{Baselines.} We establish comprehensive comparisons with four representative state-of-the-art approaches, each embodying distinct technical paradigms:
\begin{itemize}
    \item \textit{SD2}~\cite{Rombach2022SD2}. A fundamental baseline that performs independent simultaneous inpainting across all multi-view images;
    \item \textit{GridPrior+DU}~\cite{weber2024nerfiller}. An extension of the Dataset Update (DU) framework that processes random image batches in $2\times2$ grid patterns during iterative updates;
    \item \textit{NeRFiller}~\cite{weber2024nerfiller}. A progressive joint-view inpainting strategy built upon SD2, emphasizing view-consistent content generation;
    \item \textit{MVInpainter}~\cite{cao2024mvinpainter}. A reference-guided approach that fine-tunes SD2 and incorporates single-view inpainting results as reference information.
\end{itemize}
We configure \textit{GridPrior+DU} and \textit{NeRFiller} to process twelve images per batch to meet computational constraints and provide \textit{MVInpainter} with one inpainted reference image per scene. \textcolor{black}{All above methods are built upon 3DGS} and are evaluated on 3D Gaussian scenes with identical masked regions and camera poses to ensure fair comparison.

\begin{figure*}
    \centering
    \includegraphics[width=\linewidth]{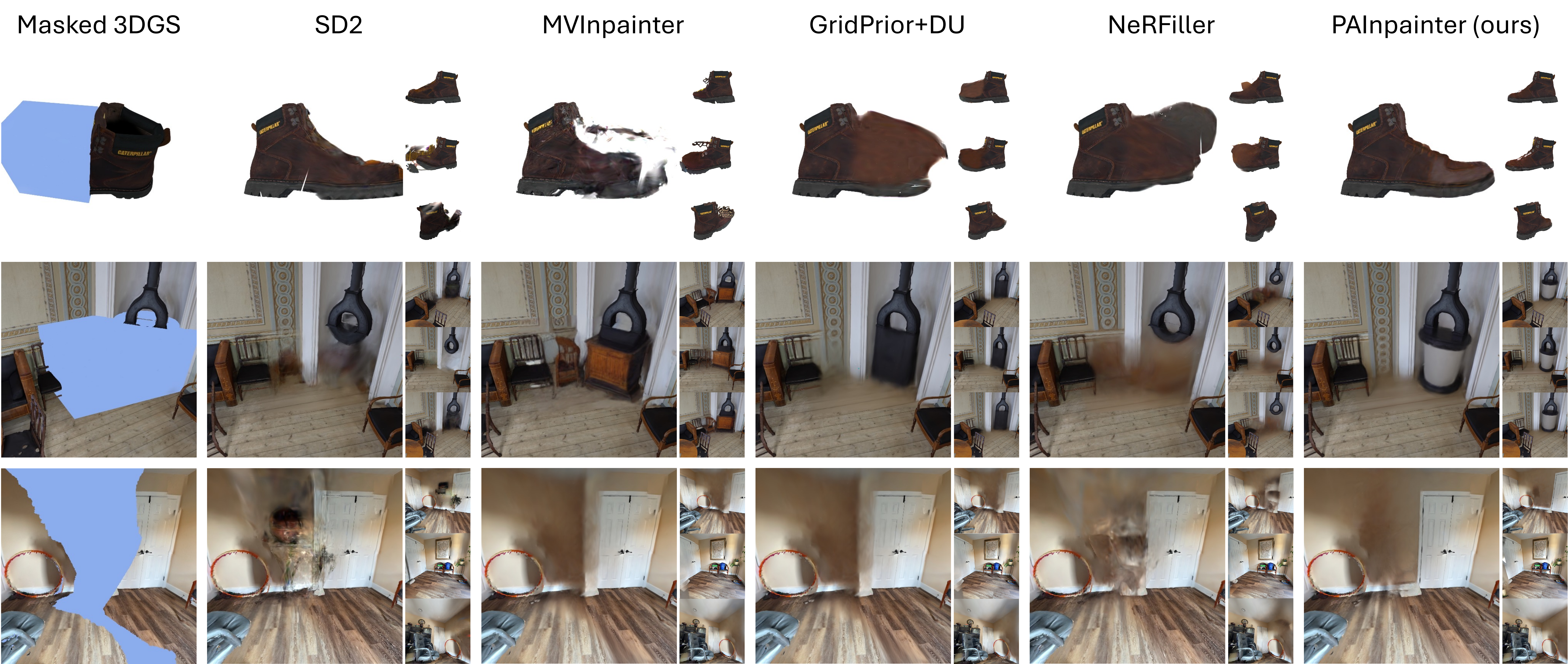}
    \caption{Qualitative comparison of 3D Gaussian inpainting results. Three scenes are shown (rows): Boot, Norway, and Office. For each scene, we present the initial masked 3D Gaussian scene (blue regions indicate missing content) and inpainting results from different methods. Four viewpoints are rendered to demonstrate multi-view consistency.}
    \label{fig:experiment}
\end{figure*}

\noindent
\textbf{Metric.} We assess the performance through quantitative analysis of rendered image quality from inpainted 3D Gaussian scenes. Our evaluation protocol employs four well-established metrics: PSNR~\cite{gonzales1987digital} for pixel-wise accuracy, SSIM~\cite{wang2004image} for structural similarity, LPIPS~\cite{zhang2018perceptual} for perceptual quality, and FID~\cite{heusel2017gans} for distribution alignment between the generated and ground truth images. For progressive methods that iteratively refine the inpainting results, we evaluate by comparing the rendered images from inpainted 3D scene against the inpainted image of each view after the last iteration. For single-round methods, we adopt the train-test split strategy on inpainted images: 80\% of images in the training set are used for optimizing the masked 3D Gaussian scene, while the remaining images as test set serve for evaluating the 3D inpainting results.

\subsection{Experimental Results and Analysis}

\textbf{Quantitative.} The experiment results of PAInpainter and state-of-the-art methods are presented in \cref{tab:quantitative_statistic}. \textcolor{black}{The \textit{Masked 3DGS} reports the rendering quality on the initial reconstructed 3D Gaussian scene with missing regions.} Our proposed PAInpainter outperforms all other methods across the evaluated metrics. Specifically, on the NeRFiller dataset, PAInpainter achieves significant improvements with PSNR of 29.51 dB, surpassing the strongest baseline (GridPrior+DU) by 2.54 dB. This demonstrates its superiority in generating high-quality 3D Gaussian inpainting results and highlights its strong generalization capability across diverse inpainting scenarios.

In contrast, \textit{SD2} inpaints images independently without considering multi-view consistency, resulting in lower-quality renderings of the inpainted 3D scene. However, due to its pretraining on large-scale image datasets, \textit{SD2} achieves competitive FID scores, suggesting its ability to generate plausible visual content without fine-tuning. \textit{MVInpainter}, originally designed for object-level and forward-facing task, struggles with general 3D inpainting scenarios, leading to unsatisfactory performance. \textit{NeRFiller} and \textit{GridPrior+DU} perform better on structural metrics due to their joint-view mechanism that incorporate cross-view priors. Nevertheless, these two approaches show limitations in perceptual quality, as reflected by higher FID scores.

Compared to previous methods, PAInpainter achieves improvement in both multi-view consistency and perceptual quality. This is particularly evident in its superior FID scores while maintaining leading performance in structural metrics, which indicates that the inpainted content generated by PAInpainter is not only coherent across views but also aligned with the original scene distribution, demonstrating its effectiveness and reliability in 3D inpainting.

\begin{table}[t]
    \centering
    \resizebox{\columnwidth}{!}{
        \begin{tabular}{c c c|c c c c}
        \toprule[1.4pt]
            \makecell{Graph \\ sampling} & \makecell{Inpaint content \\ propagation} &\makecell{Consistency \\ verification} & PSNR (dB) $\uparrow$ & SSIM $\uparrow$ & LPIPS $\downarrow$ & FID $\downarrow$\\ \midrule[1pt]
             - & - & - & 27.62 & 0.928 & 0.100 & 113.7 \\
            $\checkmark$ & - & - & 27.94 & 0.929 & 0.091 & 109.4 \\
            $\checkmark$ & $\checkmark$ & - & \underline{28.52} & 0.932 & \underline{0.083} & \underline{101.7} \\
            $\checkmark$ & - & $\checkmark$ & 28.47 & \underline{0.933} & 0.085 & 106.0 \\
            $\checkmark$ & $\checkmark$ & $\checkmark$ & \textbf{29.51} & \textbf{0.935} & \textbf{0.081} & \textbf{96.1} \\
        \bottomrule[1.1pt]
        \end{tabular}
    }
    \caption{Ablation study on NeRFiller dataset. Each row represents an ablated setting of our key components. The baseline uses basic iterative framework based on 3DGS without our proposed modules. Check marks ($\checkmark$) indicate the presence of corresponding module. Results demonstrate PAInpainter (all modules present) achieves optimal performance.}
    \vspace{-5pt}
    \label{tab:ablation_statistic}
\end{table}

\begin{figure*}[t]
    \centering
    \includegraphics[width=\linewidth]{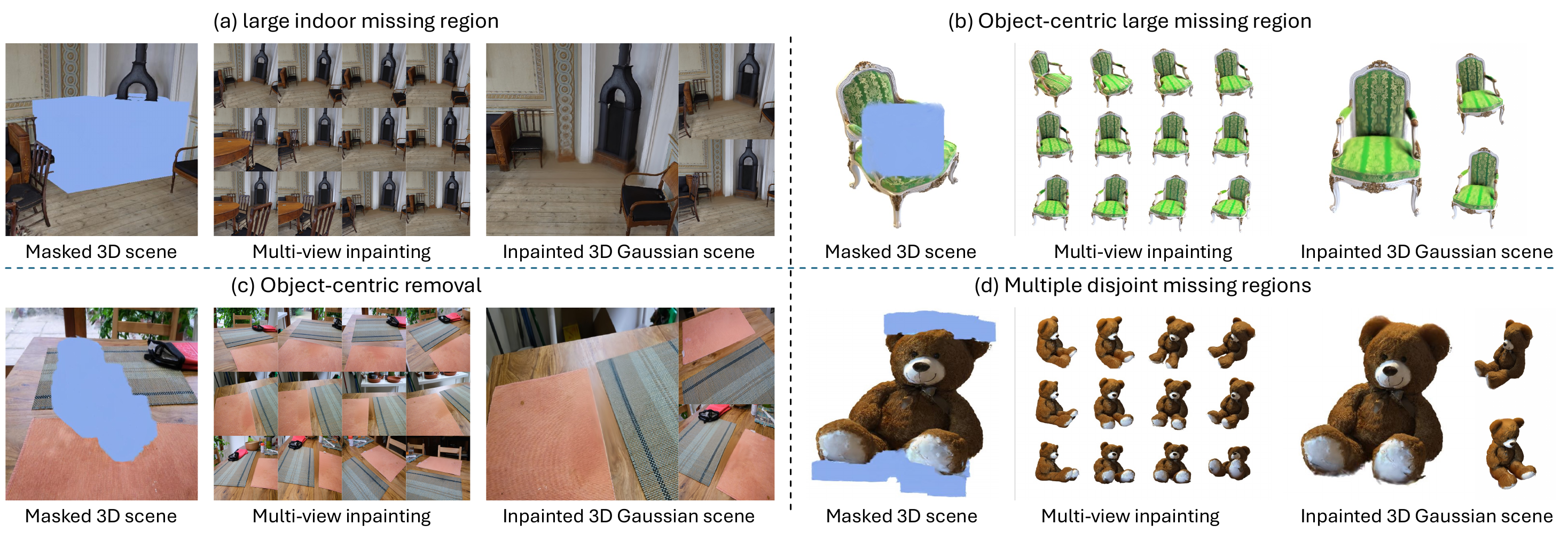}
    \caption{Visualization of PAInpainter on four representative inpainting scenarios. Each scenario shows the input, multi-view inpainting results, and the reconstructed 3D Gaussian scene, demonstrating consistent completion across varying viewpoints.}
    \label{fig:titlefigure}
\end{figure*}

\noindent\textbf{Qualitative.} Visualization comparisons are shown in \cref{fig:experiment}, where PAInpainter exhibits remarkable performance in both detail preservation and multi-view consistency. In the \enquote{boot} scene (first row), our method accurately reconstructs the intricate leather textures while maintaining geometric continuity across different viewpoints. This advantage is further evidenced in the \enquote{Norway} scene (second row), where PAInpainter faithfully recovers the fine details of the mural paintings (beside the chair) with consistent artistic style and structural integrity. Similarly, in the \enquote{office} scene (third row), our method precisely reconstructs the architectural elements of the door frame while preserving the surrounding context. Additionally, as shown in the indoor scene in \cref{fig:titlefigure}, PAInpainter showcases creativity ability by generating diverse content while maintaining scene consistency. In contrast, existing methods such as \textit{GridPrior+DU} and \textit{MVInpainter} , while achieving basic view consistency, often produce over-smoothed or distorted results, particularly in regions requiring high-fidelity detail. This comparison highlights PAInpainter's superior capability in enhancing both global structure coherence and local detail fidelity.

\begin{figure}
    \centering
    \includegraphics[width=\linewidth]{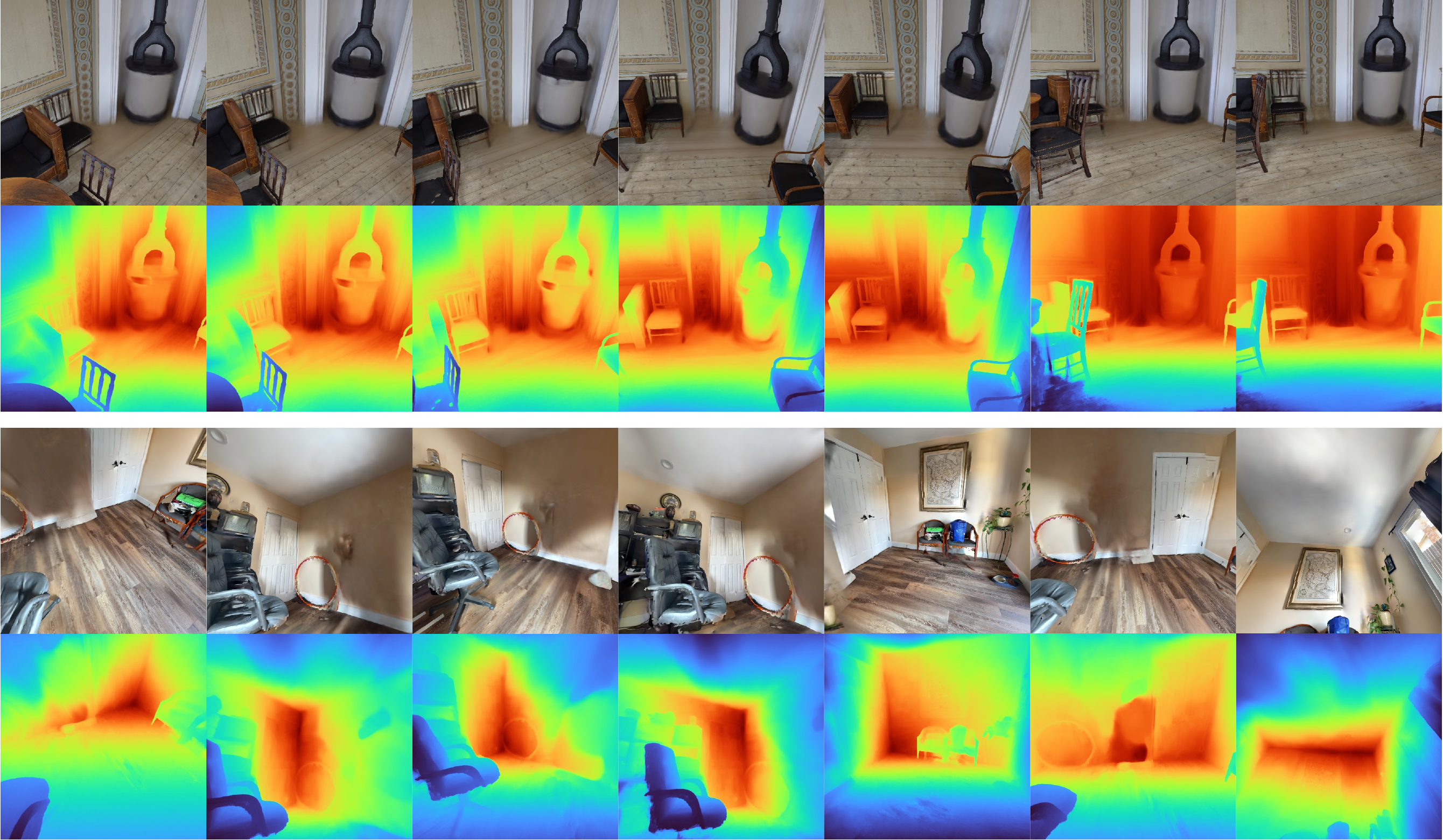}
    \caption{RGB and depth renderings of inpainted 3D Gaussian scenes from multiple viewpoints. The depth maps reveal consistent geometric reconstruction along with texture restoration.}
    \label{fig:render_depth}
\end{figure}

\subsection{Ablation Study}

We conduct ablation experiments to validate the effectiveness of each key component in PAInpainter, as shown in \cref{tab:ablation_statistic}. The baseline method adopts a basic iterative framework for multi-view image inpainting without our proposed modules. When incorporating only the graph sampling strategy, the PSNR is improved to 27.94 dB, demonstrating that reference-guided inpainting effectively promotes view consistency. Adding either inpaint content propagation or consistency verification further improves performance (PSNR: 28.52 dB and 28.47 dB, respectively), indicating both modules contribute to high-quality inpainting.

The PAInpainter equipped with all components achieves the best performance (PSNR: 29.51 dB), showing a significant improvement of 1.89 dB over the baseline. This validates our design: graph sampling provides adjacent views, content propagation enhances inpainting consistency across views, and consistency verification nominates coherent results. Notably, removing either propagation or verification module leads to similar performance degradation, suggesting these components are complementary in maintaining multi-view consistency while preserving texture details.

These ablation results corroborate our framework and previous experiments, confirming that the combination of all three proposed components is crucial for high-quality 3D Gaussian inpainting.

\subsection{Versatility \& Geometric Consistency}

We showcase diverse 3D Gaussian inpainting scenarios of PAInpainter in \cref{fig:titlefigure}, demonstrating its effectiveness across four distinct inpainting tasks. From object removal to large-area completion, our method consistently generates visually coherent results while preserving scene-specific geometric and textural details. The reconstructed 3D Gaussian scenes exhibit high fidelity across multiple viewpoints, validating the robustness of our approach in handling varying inpainting requirements.

The geometric consistency of our method is further validated through depth visualization, as shown in \cref{fig:render_depth}. Despite the absence of explicit depth supervision during 3D Gaussian optimization, the inpainted regions demonstrate naturalistic depth transitions and structural coherence with surrounding areas. The continuous depth maps demonstrate that strong multi-view consistency of PAInpainter inherently leads to accurate 3D geometry reconstruction, validating the capability in preserving both appearance and structural fidelity across different viewpoints.
\section{Conclusion}
\label{sec:concl}
In this paper, we present PAInpainter, an effective 3D Gaussian inpainter that substantially enhances multi-view consistency in 3D scene completion. Our technical contributions center on the novel perspective-aware inpainting framework, which integrates a perspective graph for adaptive view sampling, guided content propagation, and consistency verification mechanisms. Through this systematic design, PAInpainter preserves fine-grained scene details while ensuring multi-view consistency. Extensive experiments demonstrate that our method achieves superior performance across diverse inpainting scenarios.

The current PAInpainter implementation offers promising results while suggesting directions for future enhancements. The key modules in the proposed framework could be integrated into LDM in an end-to-end manner for further improved performance and deployment efficiency. Meanwhile, further exploration of sparse-view scenarios and unbounded outdoor scenes remains valuable for future work. Despite these considerations, PAInpainter demonstrates robust performance and practical utility for applications from stereo vision production to AR/VR development.
\section*{Acknowledgements}
     This research was supported by the Theme-based Research Scheme (TRS) project T45-701/22-R of the Research Grants Council (RGC), Hong Kong SAR. We thank all anonymous reviewers for their constructive feedback to improve our paper.

{
    \small
    \bibliographystyle{ieeenat_fullname}
    \bibliography{main}
}
\clearpage
\setcounter{page}{1}

\setcounter{section}{0}
\renewcommand{\thesection}{\Alph{section}}

\setcounter{figure}{0}
\renewcommand{\thefigure}{\arabic{figure}}

\setcounter{table}{0}
\renewcommand{\thetable}{\arabic{table}}

\definecolor{bl}{HTML}{2B7CCD}
\definecolor{gr}{HTML}{21A213}
\definecolor{rd}{HTML}{F66170}
\definecolor{tgr}{HTML}{E0E0E0}
\definecolor{backblue}{RGB}{210, 230, 250}
\newcommand{\default}{\cellcolor{backblue}}

\maketitlesupplementary

\section{Preliminaries}
\label{Appendix:preliminaries}

In this section, we first briefly review some preliminaries related to 3D Gaussian splatting and 2D diffusion inpainter used in PAInpainter's framework.

\noindent 
\textbf{3D Gaussian Splatting.} 3D Gaussian Splatting (3DGS) is proposed to represent 3D scenes with 3D Gaussian primitives. Given a training dataset $\mathbf{I}$ of multi-view 2D images with camera poses $\mathbf{P}$, 3DGS learns a set of colored 3D Gaussians $\mathcal{G} = \{\mathbf{g}_1, \mathbf{g}_2, \dots, \mathbf{g}_N\}$, where $N$ denotes the number of 3D Gaussians in the scene, $\mathbf{g}_i=\{\mu, \Sigma, c, \alpha \}$ and $i\in \{1, \dots, N\}$. Specifically, $\mu$ is the position where the Gaussian is centered, $\Sigma$ denotes the 3D covariance matrix, $c$ is the RGB color and $\alpha$ is the opacity attribute. Accordingly, 3DGS proposes a novel differentiable rasterization for efficient training and rendering. The rendering process can be formulated as 
\begin{equation}
    C = \qquad \sum_{i \in N} c_i\sigma_i\prod_{j=1}^{i-1}(1-\alpha_j),
\end{equation}
where $\sigma_i = \alpha_i e^{-\frac{1}{2}(x_i)^\intercal\Sigma^{-1}(x_i)}$ represents the influence of the Gaussian to the image pixel and $x_i$ is the distance between the pixel and the center of the $i$-th Gaussian. Additionally, the 3DGS training process is based on successive iterations of rendering and comparing the resulting image to the training views in $\mathbf{I}$. 

Notably, from the neural representation aspect, the 3DGS inpainting can be regarded as fine-tuning a pretrained 3DGS scene $\mathcal{G}_{u}$ with unknown region using a dataset of inpainted 2D images $\mathbf{I}_{inpainted}$.

\noindent
\textbf{2D diffusion inpainter.} 2D diffusion inpainter is a variant of Latent Diffusion Models (LDMs) focusing on inpainting masked area of 2D image~\cite{Rombach2022SD2}. In LDMs, a powerful pretrained Vector Quantised-Variational AutoEncoder (VQ-VAE) model~\cite{van2017vqvae} is employed to encode and decode the images to and from latent representations and the UNet~\cite{ronneberger2015unet} works for denoising the encoded image latent. Additionally, by introducing cross-attention layers into the UNet architecture, the generation can be controlled by text or other conditions. As a variant, the 2D diffusion inpainter expanded the UNet in LDMs to digest the mask conditioned features with unmasked area as priors and text as control condition. Thereby, the input of 2D diffusion inpainter is formulated as:
\begin{equation}
    x_t = [z_t; \mathbf{\hat{M}}; z_{\mathbf{M}}] \in \mathbb{R}^{H \times W \times 9},
\end{equation}
where $t$ indicates the time step in the diffusion; $z_{t}$ denotes the 4-channel noised latent of input image; $\mathbf{\hat{M}}$ denotes the 1-channel binary-value mask down-sampled aligned with the size of image latent; $ z_{\mathbf{M}}$ denotes the 4-channel noise-free latent feature in unmasked region. Together with encoded text prompt $y$ by the textual CLIP model~\cite{radford2021clip}, the $\mathbf{\hat{M}}$ and $ z_{\mathbf{M}}$ are concatenated as the input condition for UNet to get noise $\epsilon_{\theta}(x_t, t, y)$. The scheduler in 2D diffusion inpainter denoises the image latent in an iterative manner, and the final denoised latent is decoded to produce the inpainted image. 

\section{Implementation Details}

\subsection{Experiment Setup}
\label{Appendix:experiment_setup}
\noindent
\textbf{Method implementation.} The implementation of our 3D Gaussian Splatting (3DGS) is built upon the Nerfstudio framework. For scene initialization, we encountered a significant challenge: the large masked regions with black color in multi-view images prevent COLMAP from extracting valid 3D point clouds for 3DGS initialization. To address this, we leverage the available camera poses from the datasets and adopt different initialization strategies based on scene characteristics. For most scenes, we normalize the camera poses and randomly initialize 50k points within a unit cube to form the point cloud. However, for the scenes from SPIn-NeRF dataset, which feature uniform facet camera poses that make reconstruction from random initialization particularly challenging, we utilize their pre-computed 3D point clouds for initialization. This choice is justified by the difficulty in achieving stable reconstruction from random initialization under such camera configurations. To ensure fair comparison, all baseline methods in our experiments share identical experimental conditions, including multi-view images, camera poses, initial masked 3D Gaussian scene representations, and the optimization process of 3D Gaussians during inpainting.

\noindent
\textbf{Pretrained models.} Our framework leverages several state-of-the-art pretrained models from official repositories. For image inpainting, we adopt the "stable-diffusion-2-inpainting" model from stabilityai (Hugging Face), denoted as SD2, which serves as the primary inpainting engine for all baseline methods except MVInpainter (which employs its proprietary pretrained models). \textcolor{black}{Meanwhile, we adopt the same setting in NeRFiller, i.e. all image inpaintings are performed under the default SD2's scheduler with twenty diffusion steps.} This choice is motivated by SD2's superior performance and stability in general inpainting tasks. 

The pipeline integrates multiple specialized models for different components:
\begin{itemize}
    \item[-] \textbf{Depth Estimation}: The pretrained ZoeDepth model ("ZoeD-NK") along with off-the-shelf weights from the official torch hub, chosen for its robust depth prediction capability in diverse scenarios. \textcolor{black}{Although our framework implements the inpaint content propagation through visual projection, it remains robust against flaws caused by depth estimation under extreme condition, thanks to our iterative inpainting strategy. Additionally, the inpaint content propagation module in our framework only provides SD2 with prior information during inpainting process.To verify the reliability and generalization of ZoeDepth within our framework for 3D inpainting across various scenarios, including object-centric, indoor and outdoor scenes, we conducted comprehensive experiments on the three datasets mentioned in the main paper.}
    \item[-] \textbf{Feature Extraction}: We utilize the pretrained ResNet18 model from torchvision (default IMAGENET1K-V1 version), where we remove the last layer classification head and extract intermediate features for dual-feature consistency verification. This lightweight architecture enables efficient inference while maintaining high-quality feature representation
    \item[-] \textbf{Geometric Correspondence}: The official LoFTR model for perspective graph construction, utilized without modifications due to its proven effectiveness in establishing reliable cross-view correspondences
\end{itemize}

We maintain all models in inference mode without fine-tuning, leveraging their well-established performance as strong baselines in their respective domains. This design choice ensures reproducibility and demonstrates the generalization capability of our method. The consistent application of these models across all experimental comparisons guarantees fair evaluation.

\noindent
\textbf{Hardware Configuration and Runtime Environment.} All experiments are conducted on a server equipped with two NVIDIA RTX 3090 GPUs. We optimize the computational pipeline by dedicating one GPU to 3D Gaussian scene optimization tasks, while the other GPU handles the inference of pretrained models for image inpainting, depth estimation, and feature extraction. This parallel processing strategy significantly enhances computational efficiency while maintaining stable performance.

\subsection{Hyper-parameters Explanation}
\label{Appendix:hyper-parameters_explanation}

There are several hyper-parameters used in our PAInpainter implementation and we explain and discuss them here.

\noindent
\textbf{$\tau$ for perspective graph construction.} In our graph construction process, we employ feature matching to establish correspondences between multi-view images and utilize the average confidence score of matches to define the perspective distance between views. A higher average confidence score indicates closer perspective distance. Despite the promising performance of state-of-the-art feature matching models like LoFTR, challenging cases (e.g., significant viewpoint changes, textureless regions) may still produce unreliable matches with low confidence scores. To enhance the robustness of our graph construction method, we introduce a confidence threshold $\tau$ to filter out potentially unreliable matches. This filtering strategy effectively mitigates the impact of outliers and improves the overall stability of perspective distance estimation. We empirically set $\tau=0.4$ across all scenes in our experiments for two main reasons: 1) This value maintains a balance between match quality and quantity, ensuring sufficient valid matches for reliable perspective distance computation, and 2) It demonstrates consistent performance across diverse scenes with different viewpoint distributions and geometric complexities.While the specific choice of $\tau$ may affect individual match selection, our experiments indicate that moderate variations in the perspective graph do not significantly impact the overall inpainting performance. This robustness can be attributed to our method's inpaint content propagation strategy and consistency verification mechanism. However, for scenes with sparse viewpoint sampling or challenging viewing conditions, a lower $\tau$ value might be necessary to retain adequate matches for meaningful perspective distance estimation.

\noindent
\textbf{$k$ for adaptive adjacent images sampling.} When performing consistent multi-view inpainting, we sample $k$ adjacent images from the perspective graph for each anchor image. These sampled images form a batch for joint inpainting and subsequent 3D Gaussian optimization. In our experiments, we did not search the optimal value of $k$ and consistently set $k=8$ across all scenes to ensure fair comparison . While this parameter demonstrates robust performance in our framework, its value can be task-dependent and warrants careful consideration based on the following factors:

\begin{enumerate}
    \item \textbf{Lower Bound Constraint:} An insufficient $k$ may lead to disconnected sub-graphs during the sampling process, potentially hampering inpaint content propagation. Consider a scenario where $k=2$ and three images form a cyclic nearest neighbor relationship. This configuration necessitates additional heuristic-based anchor image selection to bridge disconnected components, introducing computational overhead and potentially compromising propagation efficiency.
    \item \textbf{Upper Bound Consideration:} Conversely, an excessive $k$ can also impact computational efficiency. As demonstrated in our findings (\cref{sec:framework}), the effectiveness of content propagation diminishes with increasing perspective distance between views. Including too many distant views in the sampling set may introduce redundant computations without contributing meaningful priors, potentially diluting the consistency of the inpainting results.
\end{enumerate}

In practical applications, the selection of $k$ should prioritize addressing the lower bound constraint to ensure connected graph components and effective content propagation. The upper bound consideration is less critical due to our consistency verification mechanism, which filters out inconsistent inpainting candidates during the refinement stage. While a larger $k$ might affect computational efficiency, it does not significantly compromise the final inpainting quality thanks to this verification safeguard. 

\noindent
\textbf{$m$ for inpainted candidates in consistency verification.} To achieve consistency verification, we need to generate multiple ($m$) inpainted candidates for each adjacent image. Thanks to our inpaint content propagation before images inpainting, most inpainted candidates are highly consistent with the anchor image. However, due to the randomness attribute of diffusion model, the consistency verification is still really important to enhance the multi-view consistency of 3D inpainting, which can be seen from our experiment results in ablation study \cref{sec:exper}. To avoid the high time consumption overhead, we set $m=4$ across all our experiments. 

\noindent
\textbf{$\eta$ for dual-feature consistency score.} In our consistency verification mechanism, we propose a weighted dual-feature consistency score that combines texture and depth features, formulated as $S = \eta S_{rgb} + (1-\eta) S_{depth}$, where $S_{rgb}$ and $S_{depth}$ represent the respective similarity scores. Through extensive experiments, we empirically set $\eta=0.7$ to prioritize fine-grained texture consistency while maintaining the benefits of geometric constraints. This weighting strategy reflects our emphasis on texture features, which directly capture the visual quality of inpainted regions, while also leveraging depth information as a valuable complementary cue. The relatively lower weight assigned to depth similarity helps mitigate potential errors introduced by the pretrained depth estimator in challenging scenes, while still providing crucial geometric constraints. This is particularly important given our use of a lightweight ResNet18 for texture feature extraction, which, while computationally efficient, may occasionally struggle to discriminate subtle texture differences under low-light conditions or in regions with repetitive patterns. In such scenarios, the depth features computed from colored depth maps demonstrate superior discriminative power, contributing significantly to the robustness of our consistency verification. Our experiments show that this balanced weighting approach provides consistent and reliable performance across diverse scenes without requiring scene-specific parameter tuning, effectively combining the strengths of both texture and geometric features while maintaining computational efficiency.

Notably, the consistent performance achieved with these empirically determined hyper-parameters ($\tau$, $k$, $m$ and $\eta$), without scene-specific tuning, underscores the robustness and practical utility of our method, making it readily applicable to real-world scenarios while maintaining its effectiveness.

\begin{algorithm}[t]
\caption{Adaptive sampling algorithm}
\label{alg:adaptive_sampling}
\begin{algorithmic}[1]
\State \textbf{Input:} perspective graph $\mathbf{G}$, adjacent hyper-parameter $k$, Iterations $iters$, threshold of consistency score $T_{s}$
\State \textbf{Initialize:} Anchor set $\mathcal{A} \gets \emptyset$, Inpainted set $\mathcal{P} \gets \emptyset$, Masked image indices set $\mathcal{I}={I_i}, i\in\{1,..., N\}$
\State Select initial anchor $I_0$ randomly from $\mathbf{G}$
\While{step $< iters\ \And\ \mathcal{I} \neq \emptyset$}
    \State $\mathcal{I}_{adj}$ $\gets$ $k$ nearest neighbors of $I_t$ from $\mathbf{G}$
    \State Update $\mathcal{A} \gets \mathcal{A} \cup \{I_t\} \cup \mathcal{I}_{adj}[:\lfloor k/2 \rfloor]$
    \State $\mathcal{I}_{adj} \gets \mathcal{I}_{adj} \cap \mathcal{I} \cup \{I_t\}$
    \If{$\mathcal{I}_{adj} \neq \emptyset$}
        \State $\mathcal{I}_{adj}' \gets$ inpainted $\mathcal{I}_{adj}$
        \State $\mathcal{S}_{adj} \gets$ consistency score of inpainted $\mathcal{I}_{adj}'$
        \State Update $\mathcal{I} \gets \mathcal{I} \setminus \mathcal{I}_{adj}'[\mathcal{S}_{adj}>T_{s}]$
        \State Update $\mathcal{P} \gets \mathcal{P} \cup \mathcal{I}_{adj}'$
        \State Optimize 3D Gaussains with $\mathcal{P}$
    \EndIf
        \State Select $I_t \gets$ random sample from $(\mathcal{I} \setminus \mathcal{A}) \cap \mathcal{P}$
\EndWhile
\While{step $< iters$}
    \State Optimize 3D Gaussains with $\mathcal{P}$
\EndWhile
\end{algorithmic}
\end{algorithm}

\subsection{Adaptive Sampling Algorithm}
\label{Appendix:adaptive_sampling_algorithm}
We formalize the adaptive sampling algorithm detailed in \cref{sec:adaptive} into pseudo-code format (\cref{alg:adaptive_sampling}) with the following key implementation details:
\begin{itemize}
    \item \textbf{State 6:} We maintain an anchor image set $\mathcal{A}$ to prevent repetitive selection of previous anchors. Additionally, the $k/2$ adjacent images of any anchor are excluded from future anchor selection to avoid local saturation in the perspective graph, ensuring comprehensive coverage of the view space.
    \item  \textbf{State 7:} The masked image set $\mathcal{I}$ adaptively tracks views requiring inpainting or refinement. Following our adaptive strategy described in \cref{sec:adaptive}, images with lower consistency scores remain in this set for subsequent refinement iterations.
    \item \textbf{State 11:} Images achieving consistency scores above the empirically determined threshold $T_s=0.9$ are removed from the masked set $\mathcal{I}$, effectively identifying well-inpainted views that require no further processing.
    \item \textbf{State 12:} An inpainted image set $\mathcal{P}$ is maintained to track all processed views throughout the algorithm's execution.
    \item \textbf{State 15:} New anchor images are selected exclusively from the inpainted set $\mathcal{P}$, excluding both previous anchors ($\mathcal{A}$) and well-inpainted views. This ensures effective propagation of high-quality inpainting results while avoiding redundant processing.
\end{itemize}

\section{Quantitative and Qualitative Results}
\label{Appendix:complementary_results}
We provide comprehensive scene-specific evaluation results to complement the average performance metrics presented in our main comparisons against state-of-the-art baseline methods. The detailed quantitative results for individual scenes are presented in \cref{tab:psnr_blender_details}, \cref{tab:psnr_spinnerf_details}  and \cref{tab:psnr_nerfiller_details} for PSNR metrics, \cref{tab:ssim_blender_details}, \cref{tab:ssim_spinnerf_details} and \cref{tab:ssim_nerfiller_details} for SSIM metrics, and \cref{tab:lpips_blender_details}, \cref{tab:lpips_spinnerf_details} and \cref{tab:lpips_nerfiller_details} for LPIPS metrics across NeRF Blender dataset, SPIn-NeRF dataset and NeRFiller dataset. In addition, we discuss the performance variation with regards to mask types and area ratios in~\cref{tab:rebuttal_masktype} (please find the examples of different mask types in main part~\cref{fig:titlefigure}), which demonstrate that performance variation is primarily influenced by mask type at reasonable ratios. PAInpainter performs better on real-world scenes and textured object scenes with more priors (SPIn-NeRF \& NeRFiller) despite larger mask ratios, compared to synthetic Blender scenes. This also reveals the 2D diffusion inpainting model's input pattern sensitivity.

\begin{figure*}[htp]
    \centering
    \includegraphics[width=\textwidth, height=0.95\textheight, keepaspectratio]{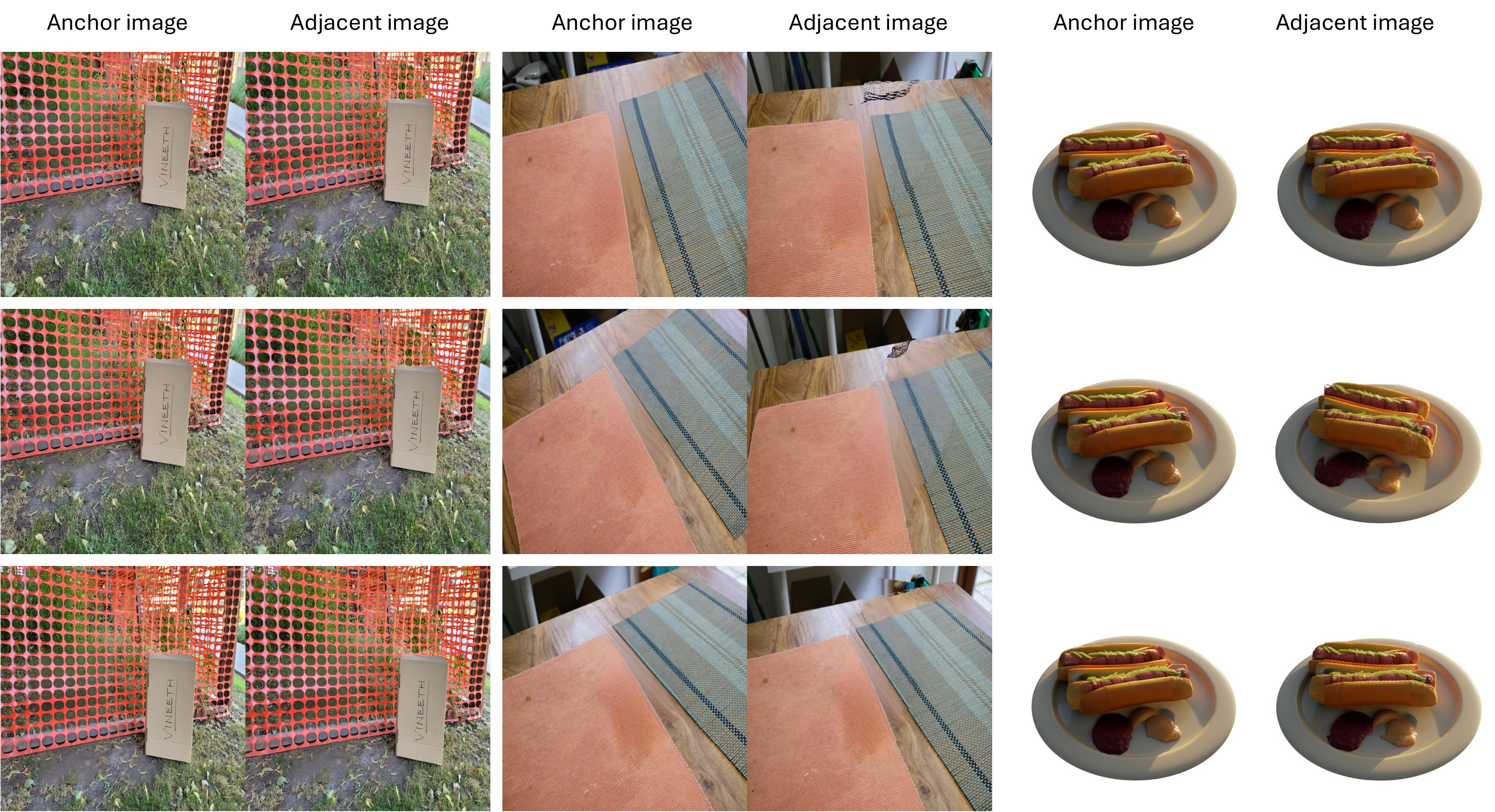}
    \caption{The inpaint content propagation between anchor images and corresponding adjacent images. With our perspective graph sampling strategy, the anchor image provides sufficient and accurate prior to adjacent images to guide consistent multi-view inpainting.}
    \label{fig:supple_projection}
\end{figure*}

\begin{figure*}[h]
    \centering
    \includegraphics[width=\linewidth]{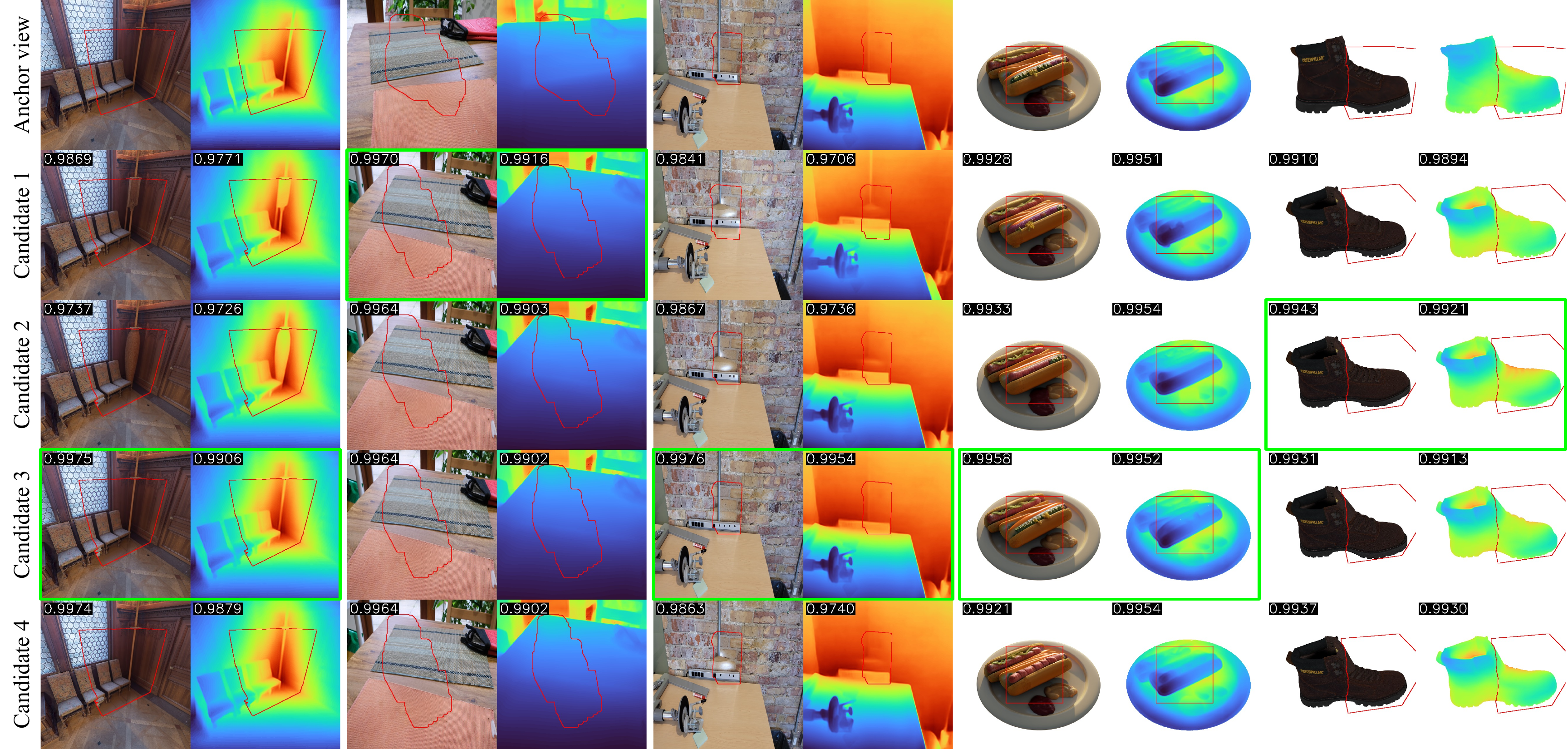}
    \caption{Visualization for consistency verification. \textcolor{rd}{Red} contours delineate mask boundaries and \textcolor{gr}{green} boxes highlight top-scoring candidates selected for 3DGS optimization. The upper-left number of each candidate represents the consistency score. This module reliably identifies inpainted regions exhibiting both textural and geometric consistency (zoom for details), enhancing performance and robustness.}
    \label{fig:verif}
\end{figure*}

\begin{figure*}[htp]
    \centering
    \includegraphics[width=\textwidth, height=0.95\textheight, keepaspectratio]{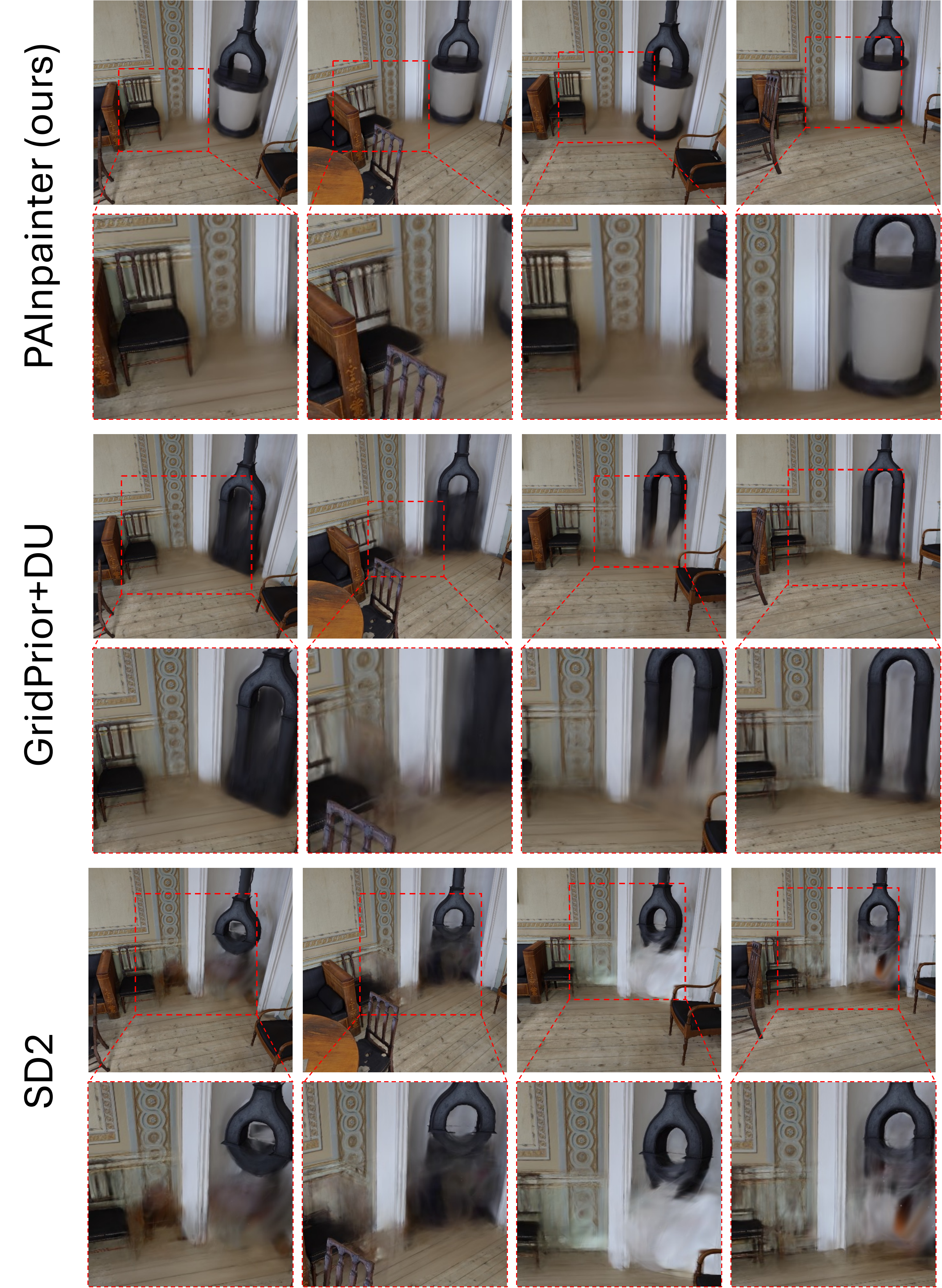}
    \caption{Details comparison in renderings of inpainted 3D scene, among PAInpainter, GridPrior+DU, SD2}
    \label{fig:zoom_norway_part1}
\end{figure*}

\begin{figure*}[htp]
    \centering
    \includegraphics[width=\textwidth, height=0.95\textheight, keepaspectratio]{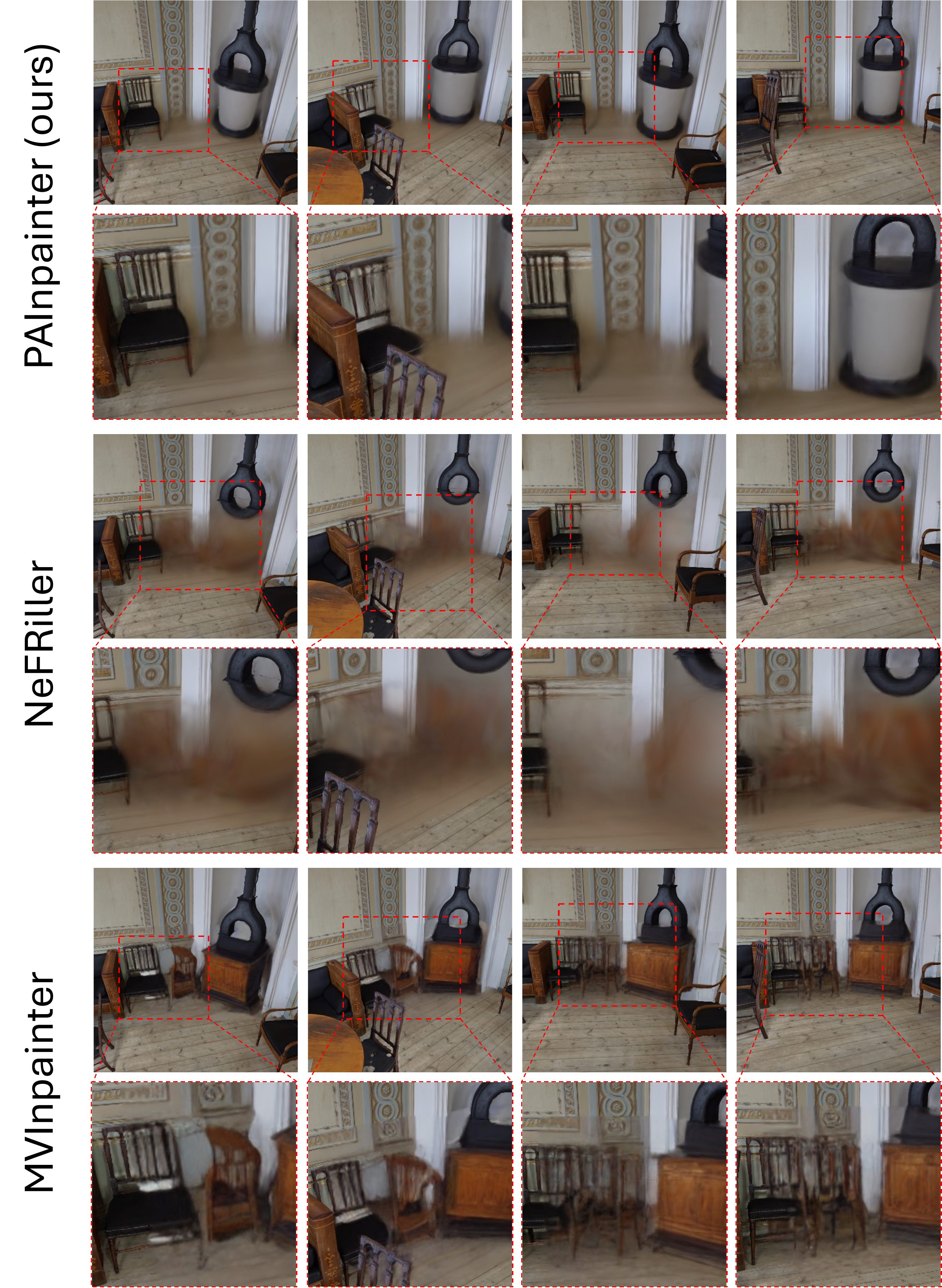}
    \caption{Details comparison in renderings of inpainted 3D scene, among PAInpainter, NeRFiller, MVInpainter}
    \label{fig:zoom_norway_part2}
\end{figure*}

We provide more supplementary qualitative results to show the details results of PAInpainter and other state-of-the-art approaches in~\cref{fig:supple_projection},~\cref{fig:verif},~\cref{fig:zoom_norway_part1} and~\cref{fig:zoom_norway_part2}.

\begin{table*}[htp]
    \centering
    \begin{tabular}{l|c c c c c c c c c }
    \toprule[1.4pt]
            & ficus & ship & lego & drums & hotdog & microphone & materials & chair & \textbf{Avg.} $\uparrow$ \\ \midrule[1pt]
        Masked 3DGS & 9.89 & 13.81 & 12.04 & 11.65 & 12.92 & 9.90 & 11.36 & 11.03 & 11.57 \\
        SD2 & 20.92 & 20.22 & 19.68 & 17.88 & 22.69 & 17.64 & \cellcolor{yellow!40} 22.14 & 22.18 & 20.42 \\
        MVinpainter & 19.56 & \cellcolor{yellow!40} 23.03 & 17.05 & 16.14 & \cellcolor{orange!40} 25.41 & 12.92 & 20.15 & 21.10 & 19.42 \\
        Grid Prior + DU & \cellcolor{yellow!40} 23.34 & 22.97 & \cellcolor{yellow!40} 21.33 & \cellcolor{yellow!40} 20.31 & \cellcolor{yellow!40} 24.95 & \cellcolor{orange!40} 22.22 & \cellcolor{orange!40} 22.17 & \cellcolor{yellow!40} 24.91 & \cellcolor{yellow!40} 22.77 \\
        NeRFiller & \cellcolor{red!40} 26.86 & \cellcolor{red!40} 24.32 & \cellcolor{red!40} 22.73 & \cellcolor{red!40} 21.63 & 24.89 & \cellcolor{yellow!40} 20.61 & 20.12 & \cellcolor{orange!40} 25.05 & \cellcolor{orange!40} 23.27 \\
        PAInpainter (ours) & \cellcolor{orange!40} 25.39 & \cellcolor{orange!40} 24.29 & \cellcolor{orange!40} 21.70 & \cellcolor{orange!40} 21.33 & \cellcolor{red!40} 26.05 & \cellcolor{red!40} 23.28 & \cellcolor{red!40} 24.84 & \cellcolor{red!40} 26.64 & \cellcolor{red!40} 24.19 \\
    \bottomrule[1.1pt]
    \end{tabular}
    \caption{PSNR 3D inpainting results for NeRF Blender dataset.}
    \label{tab:psnr_blender_details}
\end{table*}

\begin{table*}[htp]
    \centering
    \resizebox{2.05\columnwidth}{!}{
    \begin{tabular}{l|c c c c c c c c c c c}
    \toprule[1.4pt]
            & 1(bench) & 2(tree) & 3(backpack) & 4(stairs) & 7(well) & 9(wall) & 10(yard) & 12(garden) & book & trash & \textbf{Avg.} $\uparrow$ \\ \midrule[1pt]
        Masked 3DGS & 12.07 & 12.77 & 11.88 & 9.86 & 12.74 & 13.98 & 16.89 & 12.00 & 14.74 & 17.72 & 13.46\\
        SD2 & 22.68 & 24.57 & \cellcolor{orange!40} 21.69 & 25.96 & 26.82 & 21.35 & 22.08 & 21.38 & 23.42 & 24.89 & 23.48 \\
        MVinpainter & \cellcolor{yellow!40} 22.94 & 23.25 & 20.85 & \cellcolor{yellow!40} 28.15 & \cellcolor{orange!40} 28.35 & 23.41  & \cellcolor{yellow!40} 24.17 & \cellcolor{orange!40} 23.93 & 26.72 & 26.18 & 24.80 \\
        Grid Prior + DU & 22.06 & \cellcolor{yellow!40} 24.69 & 21.25 & \cellcolor{orange!40} 28.28 & \cellcolor{yellow!40} 27.89 & \cellcolor{yellow!40} 24.43 & \cellcolor{orange!40} 24.61 & 22.04 & \cellcolor{orange!40} 28.55 & \cellcolor{orange!40} 28.07 & \cellcolor{yellow!40} 25.19 \\
        NeRFiller & \cellcolor{orange!40} 23.08 & \cellcolor{orange!40} 24.74 & \cellcolor{red!40} 21.76 & 28.03 & 26.42 & \cellcolor{orange!40} 24.66 & 24.11 & \cellcolor{yellow!40} 23.72 & \cellcolor{yellow!40} 28.10 & \cellcolor{yellow!40} 27.41 & \cellcolor{orange!40} 25.20 \\
        PAInpainter (ours) & \cellcolor{red!40} 23.73 & \cellcolor{red!40} 24.93 & \cellcolor{yellow!40} 21.26 & \cellcolor{red!40} 29.39 & \cellcolor{red!40} 28.59 & \cellcolor{red!40} 25.05 & \cellcolor{red!40} 25.40 & \cellcolor{red!40} 24.31 & \cellcolor{red!40} 29.25 & \cellcolor{red!40} 28.41 & \cellcolor{red!40} 26.03 \\
        \bottomrule[1.1pt]
    \end{tabular}
    }
    \caption{PSNR 3D inpainting results for NeRF SPIn-NeRF dataset.}
    \label{tab:psnr_spinnerf_details}
\end{table*}

\begin{table*}[htp]
    \centering
    \resizebox{2.05\columnwidth}{!}{
    \begin{tabular}{l|c c c c c c c c c c c}
    \toprule[1.4pt]
            & billiards & norway & drawing & office & turtle & kitchen & bear & boot & cat & dumptruck & \textbf{Avg.} $\uparrow$ \\ \midrule[1pt]
        Masked 3DGS & 10.26 & 14.58 & 14.32 & 12.06 & 18.91 & 11.94 & 13.13 & 9.76 & 15.70 & 8.82 & 12.95 \\
        SD2 & 25.41 & 20.81 & 22.99 & 24.99 & 19.16 & 25.00 & 19.65 & \cellcolor{yellow!40} 14.72 & 16.50 & 14.37 & 20.36 \\
        MVinpainter & \cellcolor{yellow!40} 28.43 & 24.93 & 22.73 & 22.64 & 20.35 & 20.77 & \cellcolor{yellow!40} 22.24 & 14.66 & 17.73 & \cellcolor{yellow!40} 16.84 & 21.13 \\
        Grid Prior + DU & \cellcolor{red!40} 29.76 & \cellcolor{orange!40} 27.76 & \cellcolor{orange!40} 27.95 & \cellcolor{orange!40} 31.87 & \cellcolor{orange!40} 22.61 & \cellcolor{orange!40} 27.91 & \cellcolor{orange!40} 26.40 & \cellcolor{orange!40} 26.63 & \cellcolor{orange!40} 23.85 & \cellcolor{orange!40} 24.99 & \cellcolor{orange!40} 26.97 \\
        NeRFiller & 27.32 & \cellcolor{yellow!40} 25.00 & \cellcolor{yellow!40} 27.35 & \cellcolor{yellow!40} 25.75 & \cellcolor{yellow!40} 20.77 & \cellcolor{yellow!40} 25.31 & 20.90 & 13.88 & \cellcolor{yellow!40} 20.33 & 16.86 & \cellcolor{yellow!40} 22.35 \\
         PAInpainter (ours) & \cellcolor{orange!40} 29.43 & \cellcolor{red!40} 30.72 & \cellcolor{red!40} 29.26 & \cellcolor{red!40} 33.14 & \cellcolor{red!40} 30.29 & \cellcolor{red!40} 30.39 & \cellcolor{red!40} 28.33 & \cellcolor{red!40} 29.55 & \cellcolor{red!40} 26.06 & \cellcolor{red!40} 27.90 & \cellcolor{red!40} 29.51 \\
    \bottomrule[1.1pt]
\end{tabular}
    }
    \caption{PSNR 3D inpainting results for NeRFiller dataset.}
    \label{tab:psnr_nerfiller_details}
\end{table*}

\begin{table*}[htp]
    \centering
    \begin{tabular}{l|c c c c c c c c c }
    \toprule[1.4pt]
            & ficus & ship & lego & drums & hotdog & microphone & materials & chair & \textbf{Avg.} $\uparrow$ \\ \midrule[1pt]
        Masked 3DGS & 0.85 & 0.75 & 0.85 & 0.84 & 0.85 & 0.84 & 0.84 & 0.85 & 0.83 \\
        SD2 & 0.91 & 0.86 & 0.89 & 0.88 & 0.92 & 0.93 & \cellcolor{yellow!40} 0.93 & 0.91 & 0.90  \\
        MVinpainter & 0.80 & 0.82 & 0.76 & 0.74 & 0.90 & 0.79 & 0.87 & 0.84 & 0.81 \\
        Grid Prior + DU & \cellcolor{yellow!40} 0.93 & \cellcolor{yellow!40} 0.87 & \cellcolor{yellow!40} 0.90 & \cellcolor{yellow!40} 0.91 & \cellcolor{yellow!40} 0.93 & \cellcolor{red!40} 0.96 & \cellcolor{orange!40} 0.94 & \cellcolor{yellow!40}  0.93 & \cellcolor{yellow!40} 0.92 \\
        NeRFiller & \cellcolor{orange!40} 0.94 & \cellcolor{red!40} 0.88 & \cellcolor{red!40} 0.92 & \cellcolor{orange!40} 0.91 & \cellcolor{orange!40} 0.94 & \cellcolor{yellow!40} 0.93 & 0.92 & \cellcolor{orange!40} 0.93 & \cellcolor{orange!40} 0.92 \\
         PAInpainter (ours) & \cellcolor{red!40} 0.94 & \cellcolor{orange!40} 0.87 & \cellcolor{orange!40} 0.91 & \cellcolor{red!40} 0.91 & \cellcolor{red!40} 0.94 & \cellcolor{orange!40} 0.95 & \cellcolor{red!40} 0.95 & \cellcolor{red!40} 0.93 & \cellcolor{red!40} 0.92 \\
    \bottomrule[1.1pt]
\end{tabular}
    \caption{SSIM 3D inpainting results for NeRF Blender dataset.}
    \label{tab:ssim_blender_details}
\end{table*}

\begin{table*}[htp]
    \centering
    \resizebox{2.05\columnwidth}{!}{
    \begin{tabular}{l|c c c c c c c c c c c}
    \toprule[1.4pt]
            & 1(bench) & 2(tree) & 3(backpack) & 4(stairs) & 7(well) & 9(wall) & 10(yard) & 12(garden) & book & trash & \textbf{Avg.} $\uparrow$ \\ \midrule[1pt]
        Masked 3DGS & 0.28 & 0.13 & 0.31 & 0.64 & 0.50 & 0.18 & 0.50 & 0.12 & 0.71 & 0.77 & 0.41\\
        SD2 & \cellcolor{orange!40} 0.61 & \cellcolor{yellow!40} 0.72 & 0.73 & 0.83 & 0.81 & 0.54 & 0.78 & 0.65 & 0.81 & 0.80 & 0.73 \\
        MVinpainter & 0.58 & 0.67 & 0.70 & 0.83 & 0.84 & 0.64 & 0.80 & \cellcolor{orange!40} 0.81 & 0.76 & 0.81 & 0.74 \\
        Grid Prior + DU & 0.57 & 0.71 & \cellcolor{yellow!40} 0.74 & \cellcolor{orange!40} 0.87 & \cellcolor{red!40} 0.86 & \cellcolor{yellow!40} 0.68 & \cellcolor{orange!40} 0.86 & 0.78 & \cellcolor{orange!40} 0.91 & \cellcolor{orange!40} 0.89 & \cellcolor{yellow!40} 0.79 \\
        NeRFiller & \cellcolor{yellow!40} 0.60 & \cellcolor{orange!40} 0.72 & \cellcolor{red!40} 0.75 & \cellcolor{yellow!40} 0.86 & \cellcolor{yellow!40} 0.85 & \cellcolor{red!40} 0.71 & \cellcolor{yellow!40} 0.84 & \cellcolor{yellow!40} 0.80 & \cellcolor{yellow!40} 0.89 & \cellcolor{yellow!40} 0.87 & \cellcolor{orange!40} 0.79 \\
         PAInpainter (ours) & \cellcolor{red!40} 0.63 & \cellcolor{red!40} 0.75 & \cellcolor{orange!40} 0.74 & \cellcolor{red!40} 0.88 & \cellcolor{orange!40} 0.85 & \cellcolor{orange!40} 0.70 & \cellcolor{red!40} 0.89 & \cellcolor{red!40} 0.83 & \cellcolor{red!40} 0.91 & \cellcolor{red!40} 0.91 & \cellcolor{red!40} 0.81 \\
    \bottomrule[1.1pt]
\end{tabular}
    }
    \caption{SSIM 3D inpainting results for SPIn-NeRF dataset.}
    \label{tab:ssim_spinnerf_details}
\end{table*}

\begin{table*}[htp]
    \centering
    \resizebox{2.05\columnwidth}{!}{
    \begin{tabular}{l|c c c c c c c c c c c}
    \toprule[1.4pt]
            & billiards & norway & drawing & office & turtle & kitchen & bear & boot & cat & dumptruck & \textbf{Avg.} $\uparrow$ \\ \midrule[1pt]
        Masked 3DGS & 0.68 & 0.66 & 0.66 & 0.72 & 0.87 & 0.73 & 0.87 & 0.77 & 0.87 & 0.74 & 0.76 \\
        SD2 & 0.86 & 0.83 & 0.77 & 0.87 & 0.86 & 0.79 & 0.91 & 0.85 & 0.87 & 0.82 & 0.84 \\
        MVinpainter & 0.85 & 0.75 & 0.65 & 0.83 & 0.85 & 0.66 & 0.89 & 0.82 & 0.85 & 0.81 & 0.80 \\
        Grid Prior + DU & \cellcolor{orange!40} 0.92 & \cellcolor{orange!40} 0.91 & \cellcolor{yellow!40} 0.86 & \cellcolor{orange!40} 0.95 & \cellcolor{orange!40} 0.91 & \cellcolor{orange!40} 0.86 & \cellcolor{orange!40} 0.96 & \cellcolor{orange!40} 0.95 & \cellcolor{orange!40} 0.94 & \cellcolor{orange!40} 0.93 & \cellcolor{orange!40} 0.92 \\
        NeRFiller & \cellcolor{yellow!40} 0.89 & \cellcolor{yellow!40} 0.88 & \cellcolor{orange!40} 0.86 & \cellcolor{yellow!40} 0.90 & \cellcolor{yellow!40} 0.89 & \cellcolor{yellow!40} 0.80 & \cellcolor{yellow!40} 0.92 & \cellcolor{yellow!40} 0.85 & \cellcolor{yellow!40} 0.90 & \cellcolor{yellow!40} 0.87 & \cellcolor{yellow!40} 0.88 \\
         PAInpainter (ours) & \cellcolor{red!40} 0.92 & \cellcolor{red!40} 0.93 & \cellcolor{red!40} 0.88 & \cellcolor{red!40} 0.95 & \cellcolor{red!40} 0.96 & \cellcolor{red!40} 0.90 & \cellcolor{red!40} 0.96 & \cellcolor{red!40} 0.96 & \cellcolor{red!40} 0.94 & \cellcolor{red!40} 0.95 & \cellcolor{red!40} 0.94 \\
    \bottomrule[1.1pt]
\end{tabular}
    }
    \caption{SSIM 3D inpainting results for NeRFiller dataset.}
    \label{tab:ssim_nerfiller_details}
\end{table*}

\begin{table*}[htp]
    \centering
    \begin{tabular}{l|c c c c c c c c c }
    \toprule[1.4pt]
            & ficus  & ship & lego & drums & hotdog & microphone & materials & chair & \textbf{Avg.} $\downarrow$ \\ \midrule[1pt]
        Masked 3DGS & 0.21 & 0.26 & 0.17 & 0.18 & 0.19 & 0.19 & 0.17 & 0.18 & 0.19 \\
        SD2 & 0.07 & 0.13 & 0.09 & 0.11 & 0.09 & 0.09 & 0.05 & 0.08 & 0.09 \\
        MVinpainter & 0.20 & \cellcolor{orange!40} 0.12 & 0.19 & 0.21 & 0.08 & 0.35 & 0.08 & 0.15 & 0.17 \\
        Grid Prior + DU & \cellcolor{yellow!40} 0.07 & \cellcolor{red!40} 0.12 & \cellcolor{yellow!40} 0.09 & \cellcolor{yellow!40} 0.10 & \cellcolor{orange!40} 0.08 & \cellcolor{red!40} 0.05 & \cellcolor{orange!40} 0.06 & \cellcolor{orange!40} 0.07 & \cellcolor{orange!40} 0.08 \\
        NeRFiller & \cellcolor{red!40} 0.06 & 0.13 & \cellcolor{red!40} 0.08 & \cellcolor{orange!40} 0.09 & \cellcolor{yellow!40}  0.08 & \cellcolor{yellow!40} 0.08 & \cellcolor{yellow!40} 0.08 & \cellcolor{yellow!40} 0.08 & \cellcolor{yellow!40} 0.09 \\
         PAInpainter (ours) & \cellcolor{orange!40} 0.07 & \cellcolor{yellow!40} 0.13 & \cellcolor{orange!40} 0.09 & \cellcolor{red!40} 0.09 & \cellcolor{red!40} 0.07 & \cellcolor{orange!40} 0.06 & \cellcolor{red!40} 0.04 & \cellcolor{red!40} 0.06 & \cellcolor{red!40} 0.08 \\
    \bottomrule[1.1pt]
\end{tabular}
    \caption{LPIPS 3D inpainting results for NeRF Blender dataset.}
    \label{tab:lpips_blender_details}
\end{table*}

\begin{table*}[htp]
    \centering
    \resizebox{2.05\columnwidth}{!}{
    \begin{tabular}{l|c c c c c c c c c c c}
    \toprule[1.4pt]
            & 1(bench) & 2(tree) & 3(backpack) & 4(stairs) & 7(well) & 9(wall) & 10(yard) & 12(garden) & book & trash & \textbf{Avg.} $\downarrow$\\ \midrule[1pt]
        Masked 3DGS & 0.51 & 0.55 & 0.42 & 0.30 & 0.34 & 0.65 & 0.24 & 0.61 & 0.27 & 0.16 & 0.40 \\
        SD2 & \cellcolor{yellow!40} 0.37 & \cellcolor{yellow!40} 0.24 & \cellcolor{orange!40} 0.13 & 0.14 & \cellcolor{orange!40} 0.11 & 0.47 & 0.13 & 0.39 & 0.16 & 0.13 & 0.23 \\
        MVinpainter & 0.42 & 0.34 & 0.19 & \cellcolor{yellow!40} 0.11 & \cellcolor{red!40} 0.10 & 0.31 & 0.13 & \cellcolor{orange!40} 0.17 & 0.17 & 0.18 & 0.21\\
        Grid Prior + DU & 0.44 & 0.35 & 0.18 & 0.12 & 0.14 & 0.26 & 0.11 & 0.21 & \cellcolor{orange!40} 0.08 & \cellcolor{orange!40} 0.08 & \cellcolor{yellow!40} 0.20 \\
        NeRFiller & \cellcolor{orange!40} 0.37 & \cellcolor{orange!40} 0.22 & \cellcolor{red!40} 0.13 & \cellcolor{orange!40} 0.11 & 0.13 & \cellcolor{orange!40} 0.26 & \cellcolor{orange!40} 0.10 & \cellcolor{yellow!40} 0.18 & \cellcolor{yellow!40} 0.10 & \cellcolor{yellow!40} 0.09 & \cellcolor{orange!40} 0.17 \\
         PAInpainter (ours) & \cellcolor{red!40} 0.35 & \cellcolor{red!40} 0.19 & \cellcolor{yellow!40} 0.17 & \cellcolor{red!40} 0.08 & \cellcolor{yellow!40} 0.13 & \cellcolor{red!40} 0.19 & \cellcolor{red!40} 0.09 & \cellcolor{red!40} 0.15 & \cellcolor{red!40} 0.08 & \cellcolor{red!40} 0.08 & \cellcolor{red!40} 0.15 \\
    \bottomrule[1.1pt]
    \end{tabular}
    }
    \caption{LPIPS 3D inpainting results for SPIn-NeRF dataset.}
    \label{tab:lpips_spinnerf_details}
\end{table*}

\begin{table*}[htp]
    \centering
    \resizebox{2.05\columnwidth}{!}{
    \begin{tabular}{l|c c c c c c c c c c c}
        \toprule[1.4pt]
                & billiards & norway & drawing & office & turtle & kitchen & bear & boot & cat & dumptruck & \textbf{Avg.} $\downarrow$ \\ \midrule[1pt]
            Masked 3DGS & 0.33 & 0.32 & 0.33 & 0.28 & 0.21 & 0.29 & 0.19 & 0.30 & 0.21 & 0.33 & 0.28\\
            SD2 & 0.11 & 0.17 & 0.19 & 0.16 & 0.17 & 0.16 & 0.11 & \cellcolor{yellow!40} 0.21 & 0.20 & 0.26 & 0.17  \\
            MVinpainter & \cellcolor{orange!40} 0.09 & 0.17 & 0.21 & 0.18 & \cellcolor{orange!40} 0.15 & 0.26 & 0.09 & 0.24 & 0.19 & 0.24 & 0.18 \\
            Grid Prior + DU &  0.10 & \cellcolor{orange!40} 0.11 & \cellcolor{yellow!40} 0.16 & \cellcolor{orange!40} 0.09 & 0.20 & \cellcolor{orange!40} 0.15 & \cellcolor{orange!40} 0.06 & \cellcolor{orange!40} 0.10 & \cellcolor{orange!40} 0.14 & \cellcolor{orange!40} 0.15 & \cellcolor{orange!40} 0.13\\
            NeRFiller & \cellcolor{yellow!40} 0.10 & \cellcolor{yellow!40} 0.13 & \cellcolor{orange!40} 0.14 & \cellcolor{yellow!40} 0.14 & \cellcolor{yellow!40} 0.16 & \cellcolor{yellow!40} 0.15 & \cellcolor{yellow!40} 0.08 & 0.24 & \cellcolor{yellow!40} 0.15 & \cellcolor{yellow!40} 0.22 & \cellcolor{yellow!40} 0.15 \\
             PAInpainter (ours) & \cellcolor{red!40} 0.07 & \cellcolor{red!40} 0.07 & \cellcolor{red!40} 0.12 & \cellcolor{red!40} 0.08 & \cellcolor{red!40} 0.07 & \cellcolor{red!40} 0.08 & \cellcolor{red!40} 0.05 & \cellcolor{red!40} 0.06 & \cellcolor{red!40} 0.11 & \cellcolor{red!40} 0.10 & \cellcolor{red!40} 0.08 \\
        \bottomrule[1.1pt]
    \end{tabular}
    }
    \caption{LPIPS 3D inpainting results for NeRFiller dataset.}
    \label{tab:lpips_nerfiller_details}
\end{table*}

\begin{table*}[h]
    \centering
    \resizebox{0.9\linewidth}{!}{
    \renewcommand\arraystretch{0.8}
        \begin{tabular}{c | c c c c|c c c}
        \toprule[1.4pt]
             & \multicolumn{4}{c|}{Mask types} & \multicolumn{3}{c}{Avg. mask area ratios} \\ \midrule[0.5pt]
            & \makecell{object-centric \\ removal} & \makecell{large indoor \\ missing region} & \makecell{object-centric large \\ missing region} & \makecell{multiple disjoint \\ missing regions} & $\le 10\% $ & $10\% \sim 20\%$ & $20\% \sim 30\%$ \\ \midrule[1pt]
            PSNR & 26.43 & 30.61 & 25.10 & 28.23 & 26.18 & 25.59 & 29.56 \\
            SSIM & 0.817 & 0.920 & 0.931 & 0.953 & 0.823 & 0.907 & 0.924 \\
            LPIPS & 0.145 & 0.085 & 0.077 & 0.077 & 0.147 & 0.086 & 0.090 \\
            FID & 124.9 & 96.0 & 100.3 & 76.9 & 117.6 & 101.4 & 104.7 \\
        \bottomrule[1.1pt]
        \end{tabular}
    }
    \caption{Performance variation upon different mask types/ratios (All 28 scenes)}
    \label{tab:rebuttal_masktype}
\end{table*}

\end{document}